\documentclass[a4paper,fleqn]{cas-dc}

\usepackage[numbers]{natbib}
\usepackage{multirow, booktabs}
\usepackage{subcaption}
\usepackage{xcolor}
\usepackage{tabularx}
\usepackage{threeparttable}
\usepackage{stfloats}
\usepackage{amsmath}
\usepackage{adjustbox}
\usepackage{placeins}
\begin{document}
\let\WriteBookmarks\relax
\def\floatpagepagefraction{1}
\def\textpagefraction{.001}

\shorttitle{Quality and similarity indicators for video HR and RR fusion}
\shortauthors{N. Nguyen et al.}

\title [mode = title]{Quality and Similarity Indicators for Video-Based Heart Rate and Respiratory Rate Fusion: A Controlled Evaluation}

\author[1]{Nhi Nguyen}[orcid=0009-0002-2090-0746]
\cormark[1]
\ead{thi.tn.nguyen@oulu.fi}
\credit{Writing – original draft, Writing – review \& editing, Visualization, Validation, Methodology, Investigation, Formal analysis, Conceptualization, Software}

\author[1]{Le Nguyen}[orcid=0000-0001-7765-1483]
\ead{le.nguyen@oulu.fi}
\credit{Writing – original draft, Writing – review \& editing, Methodology, Investigation, Formal analysis, Conceptualization, Supervision}

\author[1]{Manuel Lage Ca\~nellas}[orcid=0000-0002-4917-340X]
\ead{manuel.lage@oulu.fi}
\credit{Writing – original draft, Visualization, Dataset, Methodology}

\author[1]{Miguel \ Bordallo\ López}[orcid=0000-0002-5707-9085]
\ead{miguel.bordallo@oulu.fi}
\credit{Writing – original draft, Writing – review \& editing, Methodology, Investigation, Conceptualization, Funding acquisition, Supervision}

\author[1]{Constantino \'{A}lvarez\ Casado}[orcid=0000-0002-3052-4759]
\ead{constantino.alvarezcasado@oulu.fi}
\credit{Writing – original draft, Writing – review \& editing, Visualization, Validation, Software, Methodology, Investigation, Formal analysis, Conceptualization, Supervision}

\affiliation[1]{organization={Center for Machine Vision and Signal Analysis (CMVS), University of Oulu},
            city={Oulu},
            country={Finland}}

\cortext[1]{Corresponding author}

\begin{abstract}
Camera-based heart rate (HR) and respiratory rate (RR) estimates vary in reliability from window to window, and a common remedy is to fuse several extraction methods with weights from reference-free signal quality indicators. However, few studies test whether these indicators identify the most accurate method in each window or outperform simpler fusion. We evaluate 11 quality indicators and 8 inter-method similarity measures for fusing 7 HR and 10 RR extractors on four public datasets (75 test subjects). Filters, indicator directions, weighting parameters and a learned ridge baseline are selected on a separate five-subject set before testing. Fusion is compared with equal averaging, a constant typical-rate predictor, the best single method selected in advance and in hindsight, and a per-window oracle with a permutation-based null, using subject-level bootstrap confidence intervals. For HR, quality-weighted fusion reduced mean absolute error by 10.9\% [5.2, 18.4] relative to averaging and by 15.6\% [8.1, 23.1] relative to the method selected in advance, and was statistically indistinguishable from the best method selected in hindsight. Weighting methods by their agreement did not improve on averaging. For RR, the indicators could not identify the more accurate estimates: sharpened quality weighting increased error ($-2.6$\% [$-4.8$, $-0.8$]), fused estimates were uncorrelated with the reference ($r \leq 0.22$), and averaging was less accurate than the typical-rate predictor ($-3.1$\% [$-6.1$, $-0.4$]). With a second respiratory reference, apparent RR gains reflected estimates moving toward its typical rate. Quality-guided fusion is worthwhile for video HR; for RR, 30-s windows improved estimates but not fusion.
\end{abstract}

\begin{highlights}
\item Choices fixed in advance; fusion tested against averaging, constant and null oracle
\item Quality weighting cut video HR error by 11\% relative to averaging
\item For RR, no indicator improved fusion in 10-s or 30-s windows
\item Apparent RR fusion gains reflected shrinkage toward the typical rate
\end{highlights}


\maketitle

\section{Introduction}\label{sec:intro}

Heart rate (HR) and respiratory rate (RR) are core vital signs. RR in particular is a sensitive early marker of clinical deterioration~\cite{zimlichman2012early,kayser2023respiratory}, yet it is the vital sign most often omitted because contact sensors are inconvenient for long-term, home or neonatal monitoring~\cite{jokinen2025understanding,wenzel2023catch}. RGB cameras offer a contactless alternative: the cardiac pulse appears as subtle skin colour changes (remote photoplethysmography, rPPG), and breathing appears as motion of the head, shoulders and chest and as modulation of the rPPG waveform~\cite{de2013Robust,Wang2017Algorithmic,Casado2023Face2PPG,boccignone2025remote}.

Video-derived signals are weak relative to noise, and their quality changes within a recording. One 10-s window may contain a clean oscillation while the next is dominated by head motion, facial expression, illumination change or compression artefacts. No single extraction method dominates across datasets~\cite{Dasari2021Evaluation,Debnath2025comprehensive}, which motivates fusing several methods and weighting each window by a reference-free estimate of its reliability: intrinsic signal quality indices (SQIs) such as spectral signal-to-noise ratio, or extrinsic agreement between methods. In contact photoplethysmography, quality-based selection reduces HR error substantially~\cite{Li2007Robust}, and several video studies report gains from quality-weighted or per-segment method selection~\cite{Bobbia2016Remote,Ghanadian2018Machine,Castellano2024machine,Ernst2020More,Li2022Enhancement}.

These reports share a gap in how fusion is evaluated. Gains are usually measured against the best single method, chosen on the same test data, without asking how much of the gain any combination would give. Three simpler mechanisms could produce the same gains, and they are rarely ruled out. First, equal averaging of several methods reduces variance without any quality information. Second, when estimates barely correlate with the reference, any method that pulls estimates toward the typical rate lowers error; we call this shrinkage toward the typical rate, and a constant predictor tests it directly. Third, an oracle that picks the estimate closest to the reference in each window reaches a low error simply because it chooses among many candidates, even when no method is more accurate in particular windows; a null oracle, computed after shuffling each method's estimates across windows, measures that effect. In addition, indicator directions, subsets and weighting parameters are often chosen with access to the test sets, and the reference rate is often computed with the same spectral estimator as the predictions, which makes the agreement look better than it is.

We therefore ask a specific question and test it against these controls: can reference-free quality indicators identify the more accurate method in each window, beyond what equal averaging or predicting the typical rate achieves? We study HR and RR side by side because they differ in signal strength and window content: a 10-s window holds 10--15 heartbeats but only 1--3 breaths. The contributions are:
\begin{itemize}
    \item A controlled evaluation of quality- and similarity-guided fusion for 7 HR and 10 RR extractors on four public datasets, against equal averaging, the median, a constant predictor, the single method chosen in advance and in hindsight, an oracle and a null oracle. All choices, from filter chains to indicator directions, weighting temperature and a learned ridge baseline, are fixed on a held-out five-subject set, and inference uses a subject-level bootstrap.
    \item An analysis of what reference-free indicators know: their usefulness within a method and across methods in the same window, with partial correlations that separate quality information from the predicted rate, and a test of inter-method agreement as an alternative reliability signal.
    \item A task-dependent answer, checked against a second respiratory reference and 30-s RR windows: quality weighting helps HR, whereas for RR the indicators' weak associations with error did not improve fusion, and apparent fusion gains reduce to shrinkage toward the typical rate.
\end{itemize}

\section{Related work}\label{sec:related}

\subsection{Video-based HR and RR estimation}

Classical rPPG methods cast pulse extraction as colour-space projection or blind source separation: GREEN uses the green channel~\cite{verkruysse2008green}, CHROM projects onto a chrominance plane~\cite{de2013Robust}, POS applies a plane orthogonal to the skin tone~\cite{Wang2016Novel}, and OMIT exploits orthogonality between pulsatile and non-pulsatile components~\cite{Casado2023Face2PPG}. Deep models such as EfficientPhys~\cite{liu2021efficientphys}, PhysFormer~\cite{zitong2022physformer}, iBVPNet~\cite{Joshi2024iBVP} and Contrast-Phys~\cite{sun2022contrastphys,sun2024contrastphys+} learn spatiotemporal representations and perform well in distribution, but often underperform classical methods on out-of-distribution data~\cite{Dasari2021Evaluation,Debnath2025comprehensive}.

Video RR draws on three cues: thoraco-abdominal or head motion tracked by landmarks or optical flow~\cite{Li2014non,Janssen2016Video,balakrishnan2013detecting,schrumpf2019exploiting,mona2023distance}, respiratory modulation of the rPPG waveform~\cite{Karlen2015Estimation}, and learned models. In controlled settings, Li et~al.~\cite{Li2014non} reported over 97\% of estimates within 1 breath per minute (brpm) of a contact sensor; 
in clinical settings, errors are higher: 2.29\,brpm in a hospital~\cite{Pediaditis2022Contactless} and 2.53\,brpm in an emergency department with audio--video fusion~\cite{Harvill2025Multimodal}. 
RR in short windows is hard to localise in frequency: a 10-s window has a nominal resolution of 0.1\,Hz (6\,brpm), so the choice of spectral estimator matters~\cite{stoica2005spectral,kumar2011spectral,nguyen2025comparative}.

\subsection{Signal quality assessment}

Contact PPG has a mature SQI literature. Li and Clifford used SQI-weighted Kalman filtering for robust HR over 6000 hours of ICU recordings~\cite{Li2007Robust}; Elgendi compared eight SQIs on 106 recordings and found skewness the most discriminative~\cite{Elgendi2016Optimal}. Skewness, kurtosis, perfusion and spectral indices recur among the most informative~\cite{elgendi2016quality,elgendi2015quality,charlton2025pi-tmcc}.

For rPPG, frequency-domain SNR is the most widely used index~\cite{Feng2015Dynamic,Bobbia2016Remote,Wang2017Robust}, and Ernst et~al.\ improved HR accuracy by up to 60\% with a decision-tree ensemble of six SQIs~\cite{Ernst2020More}. Gao et~al.\ trained an LSTM quality score whose threshold reduced HR MAE from 22.13 to 2.75\,bpm~\cite{Gao2021LSTM}, and Li et~al.\ reported a 29\% gain from quality-based segment selection~\cite{Li2022Enhancement}. The best index differs between contact and video signals: comparing eight SQIs on rPPG, Elgendi et~al.\ found $N_{\mathrm{SQI}}$ and $K_{\mathrm{SQI}}$ most informative and the perfusion index uninformative~\cite{Elgendi2024Optimal}. Respiratory signal quality has received far less attention: Romano et~al.\ proposed a morphology-based SQI for breath-by-breath RR from wearable sensors~\cite{Romano2023Signal}, and no SQI set has been validated for video RR.

\subsection{Reliability-aware fusion}

Fusion in rPPG has mostly been spatial, weighting skin regions within one method~\cite{Bobbia2016Remote,Liu2023Heart}. Method-level fusion, choosing or weighting extraction methods per segment, has been explored for HR: Ghanadian et~al.\ reduced RMSE by 27\% at rest and 42\% during movement with per-segment method selection~\cite{Ghanadian2018Machine}, and Castellano et~al.\ reported 10--20\% relative gains from a learned combination of candidate methods~\cite{Castellano2024machine}. Earlier preliminary frameworks, including our prior work~\cite{Nhi2026Quality}, suggested that mapping segment-level quality indices to estimation errors could guide adaptive signal fusion.

All of these gains are measured against a single method or against unfiltered data. To our knowledge, no study of video HR or RR fusion separates the contribution of quality information from that of simple averaging or a typical-rate prior. Oracle selection is commonly used as an upper bound, including in our prior work~\cite{Nhi2026Quality}, but a minimum over several noisy estimates is optimistic even when the methods carry no complementary information, and the oracle is rarely compared with its chance level; a null-calibrated oracle has only recently been proposed, for thermal HR~\cite{Rahman2026Oracle}. The present study supplies these controls to test whether reference-free quality indicators track window-level error.

\section{Quality and similarity framework}\label{sec:framework}

\begin{figure*}[!t]
    \centering
    \includegraphics[width=0.92\linewidth, keepaspectratio]{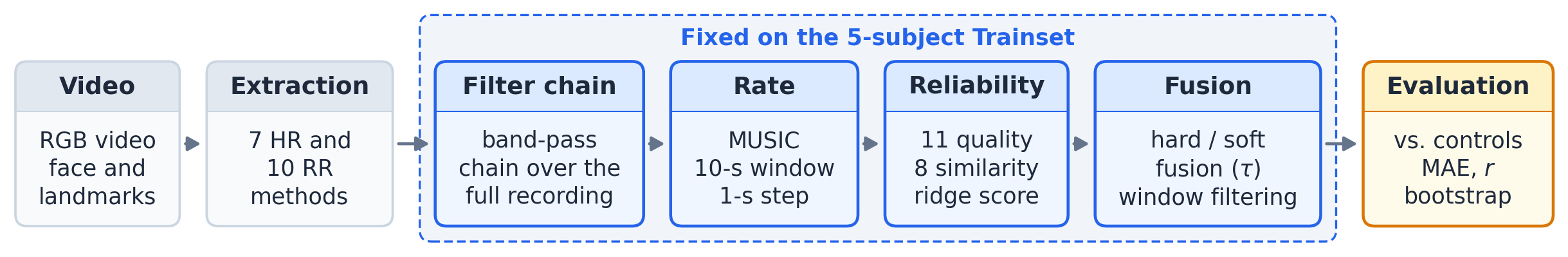}
    \caption{Quality and similarity framework. Candidate HR and RR signals are extracted by classical, motion-based and deep learning methods, filtered over the full recording, converted to rates in 10-s windows by MUSIC, scored by quality indicators, inter-method similarity or a learned ridge score, and fused or filtered. Steps in the dashed box are fixed on the five-subject Trainset before any test set is evaluated. Fusion is evaluated against a common reference and against averaging, a constant predictor and a null oracle.}
    \label{fig:pipeline}
\end{figure*}

Reliable fusion and filtering require per-window reliability scores that are informative, consistent across heterogeneous extractors and computable without access to the reference. Two sources of reliability information are defined: intrinsic signal quality indicators, computed from the signal of a single method, and inter-method similarity, which quantifies the agreement between the signals extracted by different methods from the same video as an extrinsic reliability proxy. Whether an indicator is useful is tested in two ways, since fusion requires both: the indicator should identify the windows in which a method is inaccurate (within-method), and it should rank the methods correctly within the same window (across-method). Every analysis choice is fixed on a separate five-subject set before any test set is evaluated. Figure~\ref{fig:pipeline} summarises the pipeline: signals are extracted from each video, filtered over the full recording, cut into 10-s windows with a 1-s step, converted to rates, scored by quality indicators or inter-method similarity, and fused.

\subsection{Data and signal processing}\label{sec:data}

\subsubsection{Datasets}

Four public datasets are used (Table~\ref{tab:datasets}): OMuSense-23 for model training, indicator selection and testing, and COHFACE, MAHNOB-HCI and StressID for testing only.

\textbf{OMuSense-23}~\cite{manuel2024omusense} contains recordings of 50 subjects performing guided breathing tasks in three poses, captured with an Intel RealSense D435 camera at 30\,fps ($1280\times720$ pixels), with a CONTEC CMS50E fingertip pulse oximeter and a Vernier Go Direct respiration belt as references. Subjects are split into a deep-learning training set (30 subjects), a validation set (9), an indicator selection set (5 subjects, 15 recordings; called the \emph{Trainset} and used only for the choices in Section~\ref{sec:choices}, not for training the deep models) and a test set (6 subjects, 18 recordings; \emph{OMS-Test}). All choices described in Section~\ref{sec:choices} are made on the Trainset only.

\textbf{COHFACE}~\cite{heusch2017cohface} contains 160 recordings of 39 subjects in two lighting conditions (studio and natural light) at 20\,fps, with finger blood volume pulse and a respiration belt; the respiratory reference is the rate series provided with the dataset.

\textbf{MAHNOB-HCI}~\cite{soleymani2012mahnob} contains recordings of participants watching emotion-eliciting clips at 61\,fps ($780\times580$ pixels), with ECG and a respiration belt. We use 48 sessions from 25 participants; one further session was excluded because its video is shorter than the reference.

\textbf{StressID}~\cite{Chaptoukaev2023StressID} contains recordings during stress-inducing tasks at 62\,fps. We use 60 recordings from five participants with complete video and physiological signals, with ECG and respiration references. With five subjects, StressID results are indicative.

\begin{table}[t]
\centering
\caption{Datasets. Windows are 10\,s with a 1-s step. Reference rates are mean $\pm$ SD over windows.}
\label{tab:datasets}
\small
\resizebox{\linewidth}{!}{%
\begin{tabular}{lccccccc}
\toprule
Set & Subj. & Rec. & fps & HR ref. & RR ref. & HR (bpm) & RR (brpm) \\
\midrule
Trainset & 5 & 15 & 30 & PPG & belt & 80.5 $\pm$ 11.7 & 11.9 $\pm$ 4.9 \\
OMS-Test & 6 & 18 & 30 & PPG & belt & 73.7 $\pm$ 12.9 & 12.3 $\pm$ 5.5 \\
COHFACE & 39 & 160 & 20 & PPG & rate series & 70.3 $\pm$ 11.3 & 12.7 $\pm$ 3.6 \\
MAHNOB & 25 & 48 & 61 & ECG & belt & 72.2 $\pm$ 12.3 & 15.3 $\pm$ 5.2 \\
StressID & 5 & 60 & 62 & ECG & belt & 79.8 $\pm$ 16.8 & 14.2 $\pm$ 5.6 \\
\bottomrule
\end{tabular}}
\end{table}

\subsubsection{Signal extraction}\label{sec:extraction}

Facial regions of interest of $72\times72$ pixels are defined from MediaPipe landmarks~\cite{Mediapipe2019}. For HR, seven methods are evaluated: four classical (CHROM~\cite{de2013Robust}, GREEN~\cite{verkruysse2008green}, OMIT~\cite{Casado2023Face2PPG}, POS~\cite{Wang2016Novel}) and three deep models (PhysFormer, PF~\cite{zitong2022physformer}; the self-supervised Contrast-Phys, CP~\cite{sun2022contrastphys}; and the supervised Contrast-Phys$^{+}$, CP$^{+}$~\cite{sun2024contrastphys+}). For RR, ten methods are evaluated: CHROM, GREEN and OMIT (respiratory modulation of the rPPG signal; POS is not used because its 1.6-s normalisation window attenuates the respiratory band); two motion signals from landmark distances~\cite{mona2023distance}, nose-to-lip (Mot-F) and nose-to-shoulder midpoint (Mot-S); and five deep models (EfficientPhys, EP~\cite{liu2021efficientphys}; iBVPNet, IN~\cite{Joshi2024iBVP}; PF; CP; CP$^{+}$). Deep models were trained on the 30-subject OMuSense-23 training split and validated on the 9-subject validation split~\cite{liu2023toolbox}, with the contact PPG as the target for HR and the respiration belt as the target for RR in supervised training. Depth is not used, since only OMuSense-23 provides it.

The deep models were trained at 30\,fps. Because a model trained at one frame rate sees a different number of frames per beat at another, all deep models are run at 30\,fps: MAHNOB and StressID frames are subsampled and COHFACE frames are upsampled by linear blending of neighbouring frames. Model outputs are interpolated back to the native frame rate. Clips of each model's native length are processed with 50\% overlap, standardised and cross-faded with a Hann window, which removes transients at clip boundaries.

\subsubsection{Signal filtering}\label{sec:filters}

Filters are applied to the full recording before windowing. The filter chain for each method family was selected on the Trainset from a grid of candidate chains, scored by the mean MAE rank and mean relative excess over the best candidate across methods, all four rate estimators, and both full-recording and window-wise filtering. Among candidates within one percentage point of the best, the chain with the fewest stages was used; when several had the same number of stages, the one that better preserves waveform shape was chosen (Supplement S1). The selected chains are: classical rPPG, median spike removal (0.15\,s) and an FIR band-pass (0.7--3.0\,Hz); deep rPPG, spike removal, a second-order Butterworth high-pass (0.35\,Hz) and the same FIR band-pass; classical respiratory, a second-order Butterworth band-pass (0.08--0.50\,Hz) and a Savitzky--Golay smoother (1.5\,s, order 3); deep and motion respiratory, the same Butterworth band-pass. Reference signals are filtered with fourth-order Butterworth band-passes (HR 0.7--3.0\,Hz; RR 0.08--0.50\,Hz).

\subsubsection{Rate estimation}\label{sec:estimators}

Four rate estimators were compared on the Trainset, each searching the task band ($\mathcal{B}_{\mathrm{HR}}=[0.70,3.00]$\,Hz, 42--180\,bpm; $\mathcal{B}_{\mathrm{RR}}=[0.08,0.50]$\,Hz, 4.8--30\,brpm). \textbf{FFT}: periodogram zero-padded to a 0.005-Hz grid. \textbf{Welch}: 8.5-s Hamming segments with 78\% overlap on the same grid, so that each 10-s window contains only one segment; the segment duration is identical at every frame rate. \textbf{MUSIC}: after mean removal and block-averaging to about four times the upper band edge, a Toeplitz autocorrelation matrix of size $M=\min(\max(5, N/3), 32)$ is decomposed and the pseudo-spectrum
\begin{equation}
  P_{\mathrm{MUSIC}}(f) = \Bigl(\sum_{k}\bigl|\mathbf{e}_{k}^{H}\mathbf{a}(f)\bigr|^{2}\Bigr)^{-1}
\end{equation}
is evaluated on a 0.005-Hz grid inside the band, where $\mathbf{e}_k$ are the noise-subspace eigenvectors (signal subspace of dimension 2) and $\mathbf{a}(f)$ is the steering vector. \textbf{Peak}: mean inter-peak interval within the band, with a timed zero-crossing rate when no valid interval exists. MUSIC gave the lowest mean MAE on the Trainset for both tasks (HR 7.72\,bpm vs.\ 8.33 FFT, 9.02 Welch, 9.67 Peak; RR 4.39\,brpm vs.\ 4.69, 4.92, 6.08) and is used for all methods and datasets.

\subsubsection{Reference rates}\label{sec:reference}

One common reference is used per task, computed independently of the methods' estimator. A reference computed with the same estimator as the predictions shares its frequency grid and biases, which makes errors look smaller than they are (Supplement S2.3).

\textbf{HR.} On contact PPG (OMuSense-23, COHFACE) the reference is the mean beat-to-beat interval in the window. On ECG (MAHNOB, StressID) it is the mean R-R interval from R peaks detected in the QRS band (5--15\,Hz) with a prominence threshold and a $\pm30$\% interval artefact rule; windows need at least 50\% beat coverage. The MAHNOB ground-truth signal distributed with our subset used ECG lead EXG1, which had no usable QRS complexes in 16 of 49 recordings; for each recording we used the lead with the highest median QRS prominence (EXG3 in 45, EXG2 in 4; Supplement S2.4). In one StressID recording, narrowband interference was notched before R-peak detection. Against an independent 30-s comparison rate, the beat-interval reference had an MAE of 1.17\,bpm (89.5\% of windows within 3\,bpm), whereas spectral estimates on the same contact signals deviated by 4.5--5.7\,bpm (Supplement S2.2).

\textbf{RR.} On the belt (OMuSense-23, MAHNOB, StressID), the reference rule was chosen by a check fixed before evaluation: candidate 10-s estimates were compared with breath-local rates from centred 30-s windows, with a bootstrap over recordings. FFT was closer than peak intervals by 1.79\,brpm [1.54, 2.03] over the 141 belt recordings of 41 subjects; with a median of only 1--2 breath intervals per 10-s window, peak-based rates were noisy. FFT on the belt is therefore the RR reference. MUSIC was about as close as FFT (MAE 1.28--1.85 vs.\ 1.30--2.01\,brpm across datasets; Supplement S2.2) but was not used, because the methods' estimates are computed with MUSIC. For COHFACE, the per-window mean of the provided rate series is used. Peak intervals on the belt serve as a second reference in a sensitivity analysis (Section~\ref{sec:res_robust}).

\subsubsection{Windows and metrics}

Rates are compared per 10-s window (1-s step) by the mean absolute error (MAE; bpm for HR, brpm for RR) and Pearson's $r$. Windows without a valid reference would be excluded for every method alike; none occurred after the reference corrections above. Because consecutive windows overlap by 90\%, all inference is at the subject level.

\subsection{Intrinsic quality indicators}\label{sec:indicators}

Twelve intrinsic indicators spanning temporal, morphological, spectral, statistical and complexity domains are computed per method and window (Table~\ref{tab:indicators})~\cite{vallat2023antropy,gideon2021ipr,charlton2025pi-tmcc,elgendi2016quality,elgendi2015quality}, at the sampling rate of the signal: the native frame rate (20--62\,Hz) for the classical and motion methods and 30\,Hz for the deep models.

\textbf{Morphological indicators.} For $Q_{\mathrm{TMCC}}$, each detected peak is the centre of a cycle segment $c_i$ as long as the median inter-peak interval of the window, so all segments have the same length; segments that would extend beyond the window are skipped. Each segment and the ensemble mean $\bar{c}$ are z-scored, and $Q_{\mathrm{TMCC}}$ is the mean Pearson correlation $r(c_i,\bar{c})$, set to 0 when fewer than two complete segments are found. The same procedure is applied to pulse waves for HR and to breaths for RR. The perfusion index $Q_{\mathrm{PerI}}$ is computed on the band-pass filtered signal as the median trough-to-peak amplitude divided by the median midline level.

\textbf{Spectral indicators.} For $Q_{\mathrm{SNR}}$, $Q_{\mathrm{BPR}}$ and $Q_{\mathrm{IPR}}$, the recording is filtered with the chain of Section~\ref{sec:data}, with its band-pass cutoffs widened to $[0.60, 3.50]$\,Hz for HR and $[0.07, 1.00]$\,Hz for RR so that harmonics and nearby noise remain in the spectrum, and then cut into the same windows. $Q_{\mathrm{SNR}}$ compares the mean power inside the task band $\mathcal{B}$ with the mean power outside it, and $Q_{\mathrm{BPR}}$ is the fraction of the total power inside $\mathcal{B}$; both use a Welch power spectral density, independent of the rate estimator. $Q_{\mathrm{IPR}}$ is the out-of-band fraction of the FFT magnitude spectrum. $Q_{\mathrm{BPR}}$ ranks windows exactly as $Q_{\mathrm{SNR}}$ and is excluded from fusion, leaving 11 candidates.

\textbf{Statistical and complexity indicators.} $Q_{\mathrm{KURT}}$ and $Q_{\mathrm{SKEW}}$ are the absolute excess kurtosis and skewness of the window, i.e.\ the distance from a symmetric, Gaussian-like shape; as for every indicator, whether higher values mean better or worse quality is determined on the Trainset rather than assumed. $Q_{\mathrm{HFD}}$ is estimated from the scaling of the curve length $L(k)$ with the time step $k$, and $Q_{\mathrm{KFD}}$ relates the total path length of the signal to its maximum excursion.

\begin{table*}[t]
\centering
\caption{Quality indicators and the directions inferred on the Trainset ($\downarrow$ lower is better, $\uparrow$ higher is better, -- dropped because the sign was not the same in all five Trainset subjects). $\rho_{\times}$ is the cross-method rank correlation with absolute error on the Trainset (raw sign; negative means higher values go with lower error).}
\label{tab:indicators}
\small
\resizebox{\linewidth}{!}{%
\begin{tabular}{llccrcr}
\toprule
 & & \multicolumn{2}{c}{HR} & & \multicolumn{2}{c}{RR} \\
\cmidrule{3-4}\cmidrule{6-7}
Indicator & Definition & Dir. & $\rho_{\times}$ & & Dir. & $\rho_{\times}$ \\
\midrule
Zero-crossing rate, $Q_{\mathrm{ZCR}}$ & sign changes per sample & -- & 0.11 & & -- & 0.00 \\
Hjorth mobility, $Q_{\mathrm{HJ\text{-}M}}$ & $\sqrt{\mathrm{Var}(\Delta x)/\mathrm{Var}(x)}$ & $\downarrow$ & 0.15 & & -- & $-0.01$ \\
Hjorth complexity, $Q_{\mathrm{HJ\text{-}C}}$ & mobility of $\Delta x$ / mobility of $x$ & $\downarrow$ & 0.29 & & -- & 0.05 \\
Signal-to-noise ratio, $Q_{\mathrm{SNR}}$ & mean in-band / mean out-of-band power (dB) & $\uparrow$ & $-0.25$ & & $\uparrow$ & $-0.12$ \\
Irrelevant power ratio, $Q_{\mathrm{IPR}}$ & out-of-band / total magnitude & -- & 0.04 & & -- & 0.01 \\
Kurtosis, $Q_{\mathrm{KURT}}$ & $|$excess kurtosis$|$ & -- & $-0.12$ & & $\uparrow$ & $-0.07$ \\
Skewness, $Q_{\mathrm{SKEW}}$ & $|$skewness$|$ & -- & 0.18 & & $\downarrow$ & 0.05 \\
Perfusion index, $Q_{\mathrm{PerI}}$ & median amplitude / median midline level & -- & $-0.10$ & & -- & $-0.05$ \\
Template-matching cross-corr., $Q_{\mathrm{TMCC}}$ & mean $r(c_i,\bar{c})$ over cycles & $\uparrow$ & $-0.27$ & & -- & 0.03 \\
Higuchi fractal dim., $Q_{\mathrm{HFD}}$ & slope of $\log L(k)$ vs.\ $\log(1/k)$ & $\downarrow$ & 0.23 & & $\downarrow$ & 0.03 \\
Katz fractal dim., $Q_{\mathrm{KFD}}$ & $\log_{10}(L/a)/\log_{10}(d/a)$ & -- & $-0.12$ & & -- & $-0.04$ \\
\midrule
Band power ratio, $Q_{\mathrm{BPR}}$ & in-band / total power & \multicolumn{5}{l}{ranks windows exactly as $Q_{\mathrm{SNR}}$; excluded from fusion} \\
\bottomrule
\end{tabular}}
\end{table*}

\subsection{Inter-method similarity}\label{sec:similarity}

The intrinsic indicators can only detect inconsistency with the expected properties of a physiological oscillation. They cannot detect a signal that is well formed but follows the wrong event, such as a regular head movement. Eight measures therefore quantify how well the signals $x$ and $y$ of two methods agree in a window, in three domains.

\textbf{Time domain.} Pearson correlation $S_{\rho}$, cosine similarity $S_{\cos}$, maximum normalised cross-correlation over lags $S_{\mathrm{NCC}}$, and $S_{\mathrm{MAE}} = 1/(1+\mathrm{MAE}(x,y))$.

\textbf{Frequency domain.} Negative peak-frequency difference $S_{\mathrm{PFD}} = -|f^{*}_{x}-f^{*}_{y}|$ and band-power correlation $S_{\mathrm{BPC}}$, the correlation of the two in-band spectra.

\textbf{Time--frequency.} Correlation of STFT magnitudes $S_{\mathrm{STFT}}$ and of in-band energy envelopes $S_{\mathrm{BEC}}$.

Three domains are used because no single measure is expected to be discriminative in every case: time-domain measures are sensitive to amplitude and phase, frequency-domain measures target agreement on the rate but are noisy in short windows, and time--frequency measures lie between the two.

The similarity score of method $i$ in a window is its mean similarity to the other $K-1$ methods, where $K$ is the number of methods,
\begin{equation}
  \mathrm{sim}_{i} = \frac{1}{K-1}\sum_{j\neq i} S_{ij},
\end{equation}
and higher agreement is taken as higher reliability.

\subsection{Fusion, Trainset choices and evaluation}\label{sec:choices}\label{sec:trainset}\label{sec:fusion}

\textbf{Aggregated scores.} In each window, each oriented indicator is min--max normalised across methods so that 0 marks the method that looks most reliable. The all-metric mean (AMM) averages the indicator pool; the subset mean (SMM) averages a subset chosen by exhaustive search on the Trainset. Similarity scores are oriented and normalised the same way and enter fusion exactly as quality scores do (AMM over all eight measures, or a Trainset-selected subset).

\textbf{Hard and soft fusion.} Hard fusion takes the estimate of the method with the lowest score, $\hat{y} = \hat{y}_{i^*}$, $i^*=\arg\min_i s_i$. Soft fusion weights all methods,
\begin{equation}
  w_i = \frac{\exp\bigl(-(s_i - \min_j s_j)/\tau\bigr)}{\sum_k \exp\bigl(-(s_k - \min_j s_j)/\tau\bigr)}, \qquad \hat{y} = \sum_i w_i \hat{y}_i ,
\end{equation}
with $\tau = 1$ as the untuned setting and $\tau^{*}$ as the Trainset-tuned one; $\tau=\infty$ gives the equal mean. Configurations are named by score ($Q$ quality, $S$ similarity, L learned), mode (H hard, S soft) and aggregation (AMM, Trainset-SMM).

\textbf{Learned error score.} A ridge regression predicts $\log(1+|e|)$ from the standardised pool indicators (L-Q) or from the indicators and the predicted rate (L-Q+rate), fitted on all Trainset methods and windows. The predicted error serves as the score.

\textbf{Choices fixed on the Trainset.} The filter chains, the rate estimator, the indicator directions, the SMM subsets, the ridge models and $\tau^{*}$ are all fixed on the Trainset and saved before any test set is evaluated (Supplement S3). For each indicator and task, the rank correlation between indicator value and absolute error is computed across methods within each window and averaged per subject. An indicator is kept, with the direction given by its sign, only if the sign is the same in all five subjects; otherwise it is dropped. This yields an HR pool of five indicators ($Q_{\mathrm{HJ\text{-}M}}$, $Q_{\mathrm{HJ\text{-}C}}$, $Q_{\mathrm{SNR}}$, $Q_{\mathrm{TMCC}}$, $Q_{\mathrm{HFD}}$) and an RR pool of four ($Q_{\mathrm{SNR}}$, $Q_{\mathrm{KURT}}$, $Q_{\mathrm{SKEW}}$, $Q_{\mathrm{HFD}}$). The SMM subsets were $Q_{\mathrm{HJ\text{-}C}}$ and $Q_{\mathrm{TMCC}}$ for HR and $Q_{\mathrm{SNR}}$ and $Q_{\mathrm{SKEW}}$ for RR soft fusion. The ridge regularisation strength $\alpha \in \{0.1, 1, 10, 100, 1000\}$ and the temperature of the learned score are chosen by leave-one-subject-out validation. The temperature $\tau$ is tuned over $\{0.05$, 0.1, 0.2, 0.3, 0.5, 0.75, 1, 1.5, 2, 3, 5, 10, 20, $\infty\}$ (HR $\tau^{*}=0.2$, RR $\tau^{*}=0.1$ for the ALL group; tuning curves in Supplement S4). Results are reported for all methods (ALL); groups of classical methods, deep methods, and the best classical plus the best deep method on the Trainset (DN1: POS + PF for HR; Mot-S + PF for RR) are in Supplement S7.

\textbf{Evaluation.} Fusion is compared with the best single method on each test set (\emph{Baseline}, chosen in hindsight and therefore conservative toward fusion), the best single method on the Trainset (\emph{Baseline-train}; HR: PF, RR: Mot-S), the choice a practitioner would make in advance, and the \emph{Equal mean} and \emph{Median}, which combine all methods without quality information. For each dataset, MAE is computed over all windows of that dataset, and the difference between two configurations is the relative gain $g=(\mathrm{MAE}_A-\mathrm{MAE}_B)/\mathrm{MAE}_A$, so that positive values favour $B$. The pooled gain is the unweighted mean of the four per-dataset gains. Datasets are weighted equally because they differ in recording conditions, and subject weighting would let COHFACE (39 of 75 subjects) decide the pooled result. The drawback is that StressID, with only five subjects, carries a quarter of the pooled weight. Confidence intervals (95\%, percentile) come from 2000 bootstrap resamples in which subjects are resampled within each dataset. Per-dataset intervals are given in Supplement S8, together with a subject-weighted pooled gain, $1-\sum_s \mathrm{MAE}_B(s)/\sum_s \mathrm{MAE}_A(s)$ over the 75 subjects, with each subject's MAE taken over its own windows.

\textbf{Indicator--error association.} The association between each oriented indicator and the absolute error is measured with the Spearman correlation (i) within each method across its windows (within-method $\rho$), and (ii) across methods within each window, averaged per subject (cross-method $\rho$). Positive values mean the indicator is useful. A partial correlation controlling for the predicted rate tests whether an indicator carries information beyond the rate itself. Inference uses one $\rho$ per subject and method, a two-sided Wilcoxon signed-rank test across subjects and Benjamini--Hochberg correction over all indicator--method--dataset cells of a task. The smallest attainable two-sided $p$ is 0.0625 with five subjects and 0.031 with six, so after correction significance is reachable only on COHFACE (39 subjects) and MAHNOB (25); on the smaller sets the subject-level $\rho$ values are reported descriptively.

\textbf{Window filtering.} Windows are removed in order of decreasing AMM score (0--50\%) for each method, and the resulting MAE is compared with removing windows at random (50 draws) and with removing them by true error (oracle).

\section{Results}\label{sec:results}

\subsection{Single-method HR and RR estimation}\label{sec:res_bench}

Figure~\ref{fig:heatmaps} shows single-method MAE for all estimators (full values in Supplement S5). With MUSIC, the best HR methods were CP on OMS-Test (5.46\,bpm, $r=0.70$) and MAHNOB (5.99, $r=0.62$), CP$^{+}$ on COHFACE (7.29, $r=0.51$) and PF on StressID (13.07, $r=0.06$). For RR, the best methods were PF on OMS-Test (4.55\,brpm), EP on COHFACE (3.77), and IN on MAHNOB and StressID (5.24 and 5.27); no RR method exceeded $r=0.19$ on any test set. The best method therefore differs between datasets for both tasks.

\begin{figure*}[!t]
    \centering
    \begin{subfigure}{\linewidth}
        \centering
        \includegraphics[width=\linewidth, keepaspectratio]{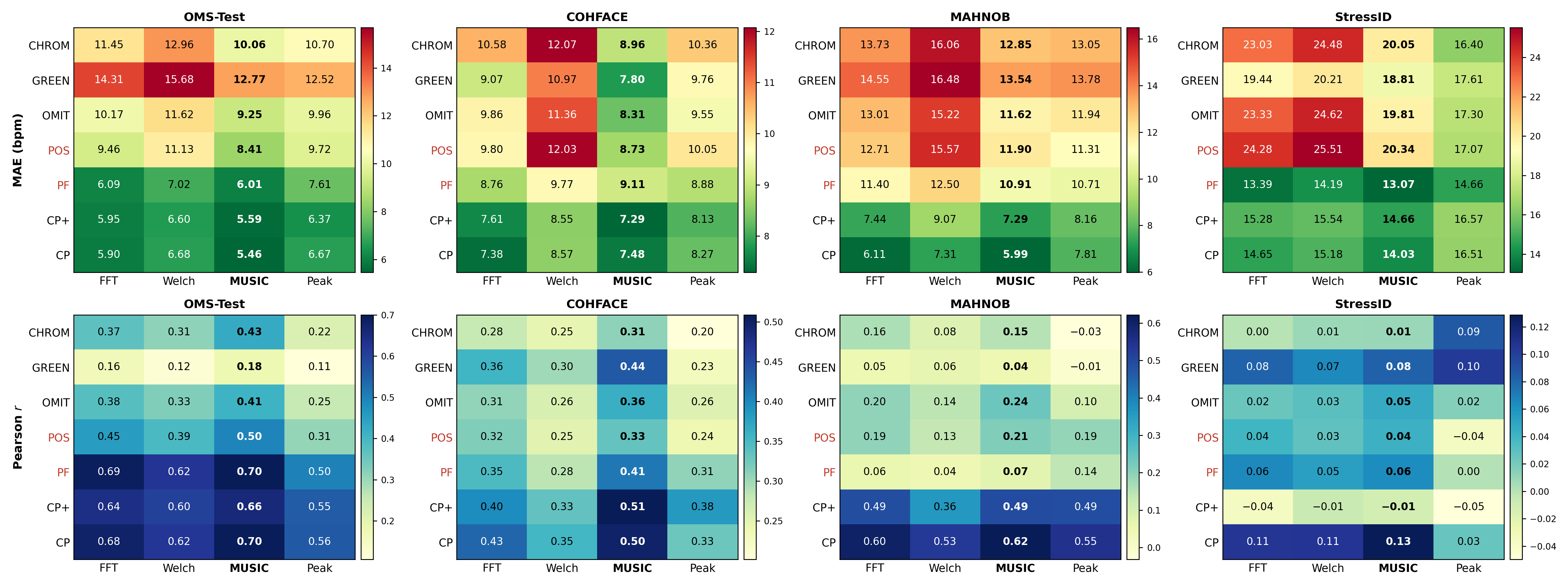}
        \caption{HR}
    \end{subfigure}
    \begin{subfigure}{\linewidth}
        \centering
        \includegraphics[width=\linewidth, keepaspectratio]{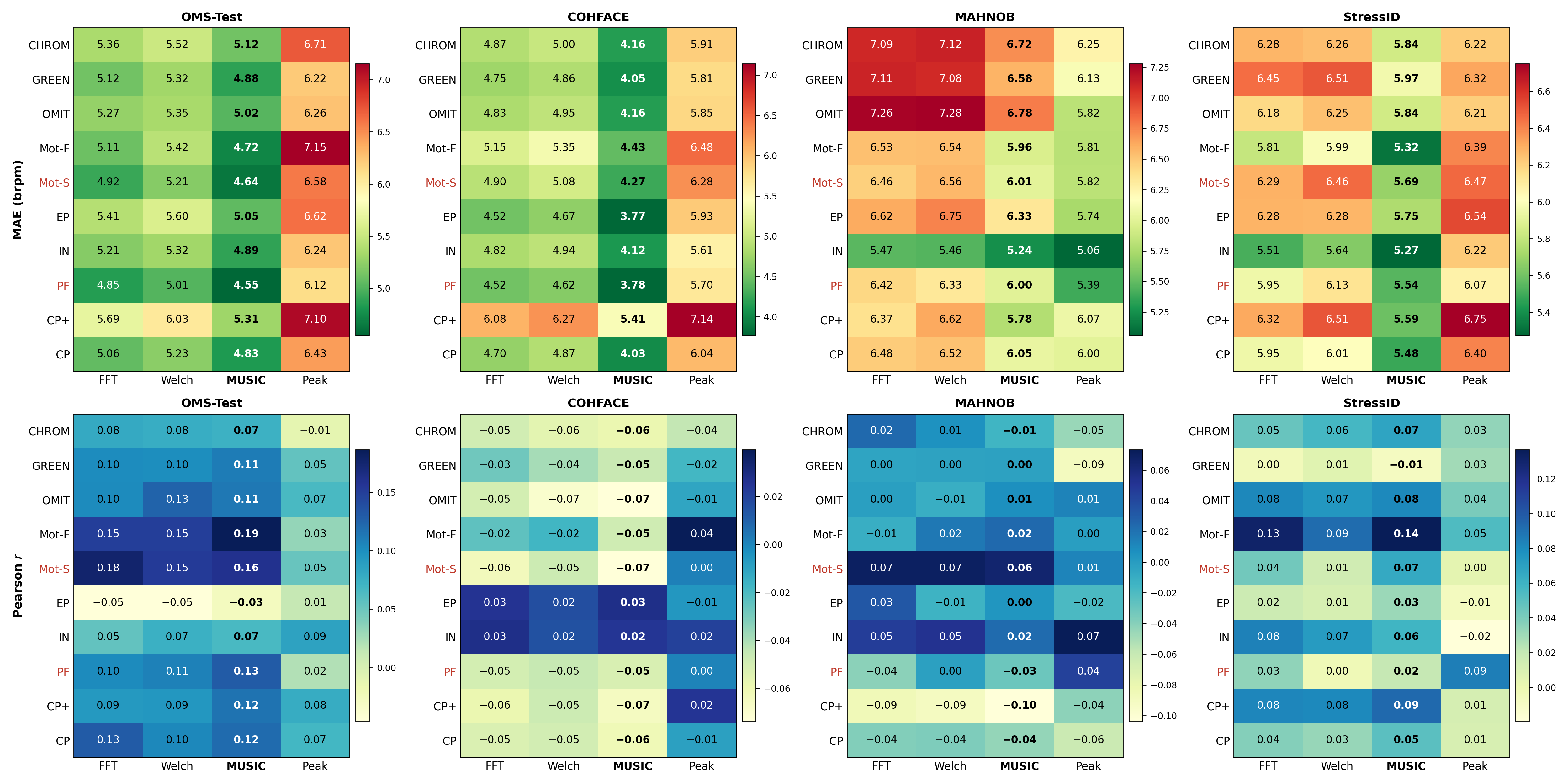}
        \caption{RR}
    \end{subfigure}
    \caption{Single-method MAE (top rows) and Pearson $r$ (bottom rows) for each method and rate estimator. The Trainset-chosen estimator (MUSIC, bold) is used for everything that follows; red labels mark the Trainset DN1 pair (HR: POS + PF; RR: Mot-S + PF).}
    \label{fig:heatmaps}
\end{figure*}

\begin{figure*}[!t]
    \centering
    \includegraphics[width=\linewidth, keepaspectratio]{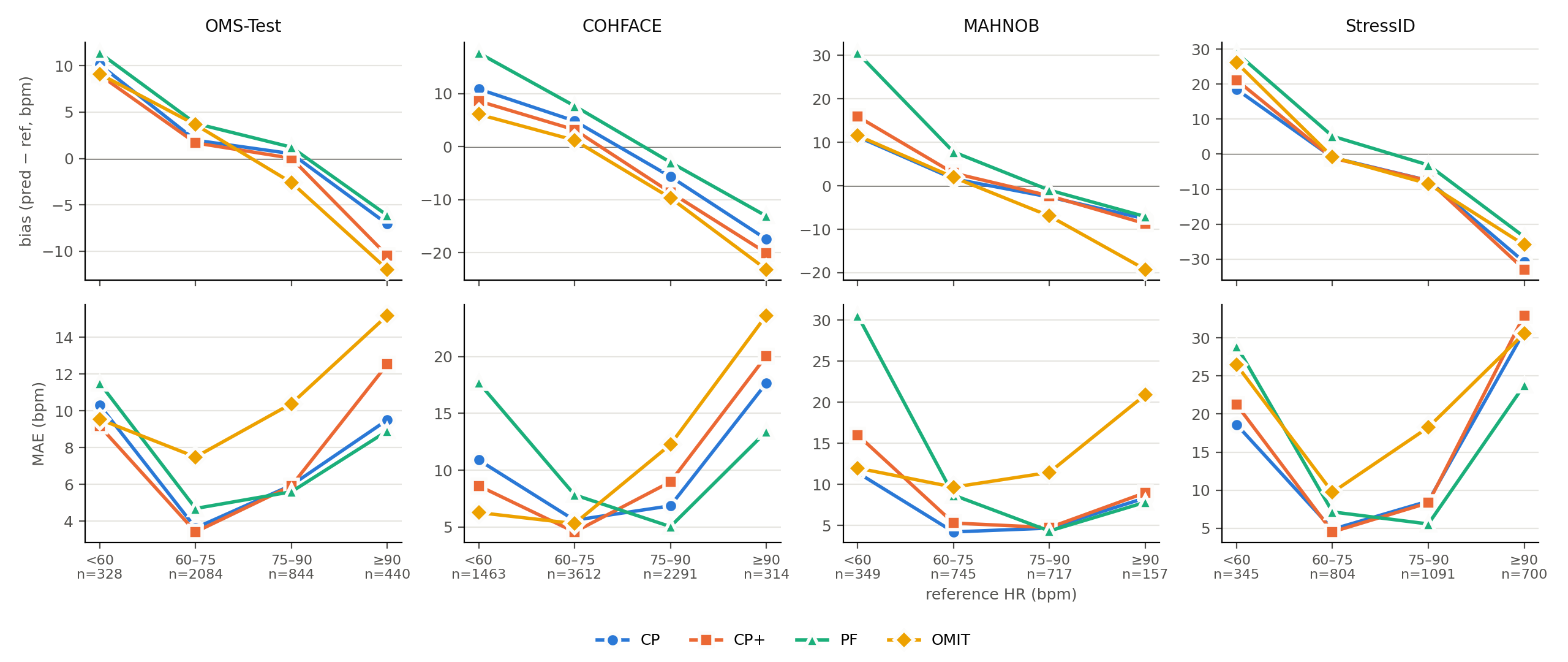}
    \caption{HR bias (top) and MAE (bottom) by reference HR for the three deep models and OMIT. All methods over-estimate low and under-estimate high heart rates.}
    \label{fig:hrdep}
\end{figure*}

HR errors depend strongly on the true rate (Fig.~\ref{fig:hrdep}). Across methods and datasets, estimates were biased upward by 6--34\,bpm when the reference was below 60\,bpm and downward by 6--33\,bpm when it was 90\,bpm or above: estimates are compressed toward 70--80\,bpm. StressID, where 24\% of windows exceed 90\,bpm during stress tasks, is most affected; there the constant predictor (13.13\,bpm) matches the best single method (13.07).

\subsection{Quality indicators and estimation error}\label{sec:res_train}\label{sec:res_within}\label{sec:res_cross}

\subsubsection{Indicator directions}

Table~\ref{tab:indicators} lists the inferred directions. Only $Q_{\mathrm{SNR}}$ (higher is better) and $Q_{\mathrm{HFD}}$ (lower is better) were kept with the same direction for both tasks. Directions inferred without any reference, using the median of the other methods as a pseudo-reference, recovered all five HR pool directions but only two of the four RR directions (Supplement S3). For RR, larger absolute kurtosis went with lower error, the opposite of the assumption that a near-Gaussian shape indicates quality; for HR, its sign was not consistent across the Trainset subjects.

\subsubsection{Within-method association}

\begin{figure*}[!t]
    \centering
    \includegraphics[width=\linewidth, keepaspectratio]{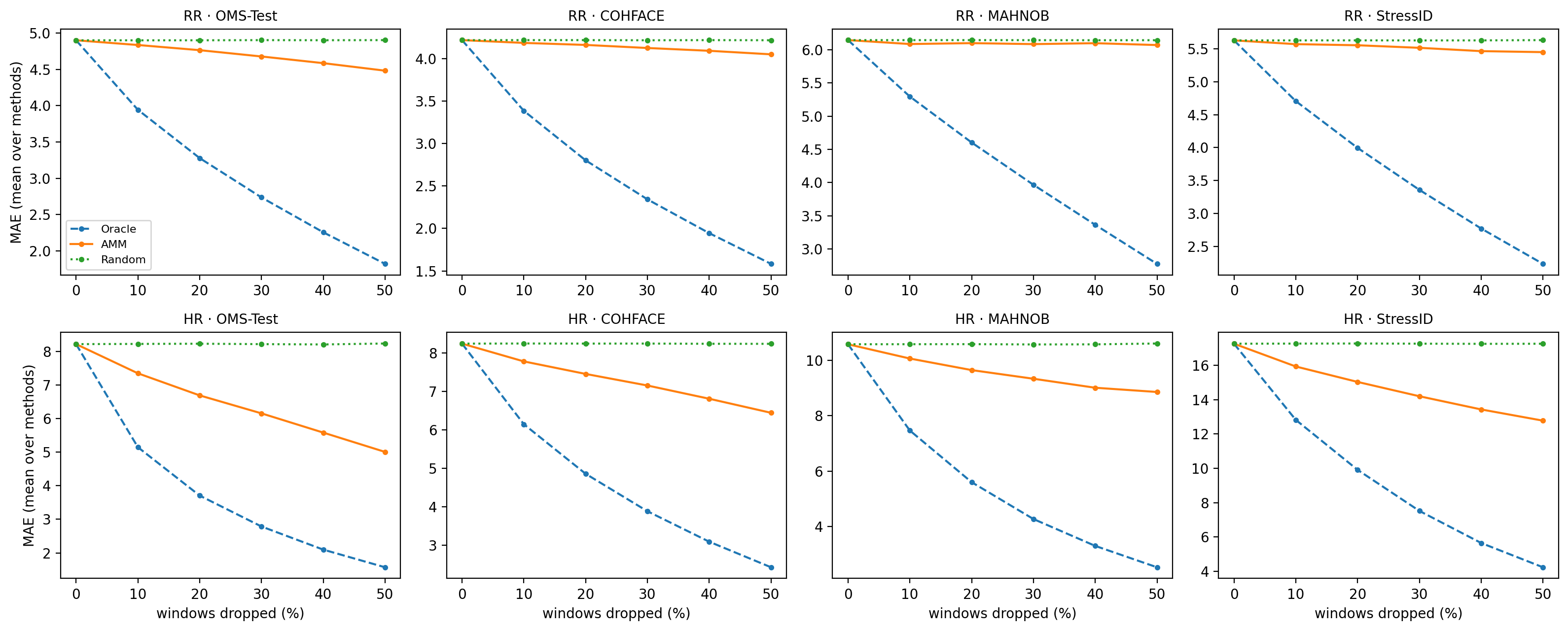}
    \caption{Filtering: MAE averaged over methods as the fraction of removed windows grows, removing windows by AMM score, at random, or by true error (oracle).}
    \label{fig:filtering}
\end{figure*}

For HR, the pool indicators correlated with a method's own error: median within-method $\rho$ for $Q_{\mathrm{HFD}}$ was 0.10--0.29 on the test sets and for $Q_{\mathrm{SNR}}$ up to 0.26 (OMS-Test). Partial correlations given the predicted rate stayed close to the raw values (e.g.\ $Q_{\mathrm{HFD}}$ 0.15--0.28), so the information is not merely the rate. Removing the 30\% of windows with the worst AMM score lowered HR MAE by 11.7--25.1\% relative to random removal (Fig.~\ref{fig:filtering}). On COHFACE and MAHNOB, where tests have power, the within-method correlation survived Benjamini--Hochberg correction in the useful direction for $Q_{\mathrm{SNR}}$ and $Q_{\mathrm{HFD}}$ in 3 of 14 method--dataset cells each, and no pool indicator was significant in the harmful direction. Given the predicted rate, most pool indicators reached significance in more cells, not fewer ($Q_{\mathrm{HFD}}$ 11 of 14, $Q_{\mathrm{HJ\text{-}M}}$ 9, $Q_{\mathrm{TMCC}}$ 6), so the HR information is not a by-product of the estimated rate; $Q_{\mathrm{SNR}}$ fell from 3 cells to 1 and $Q_{\mathrm{HJ\text{-}C}}$ split (3 useful, 3 harmful; Supplement S6).

\begin{figure}
    \centering
    \includegraphics[
        width=0.95\columnwidth,
        keepaspectratio
    ]{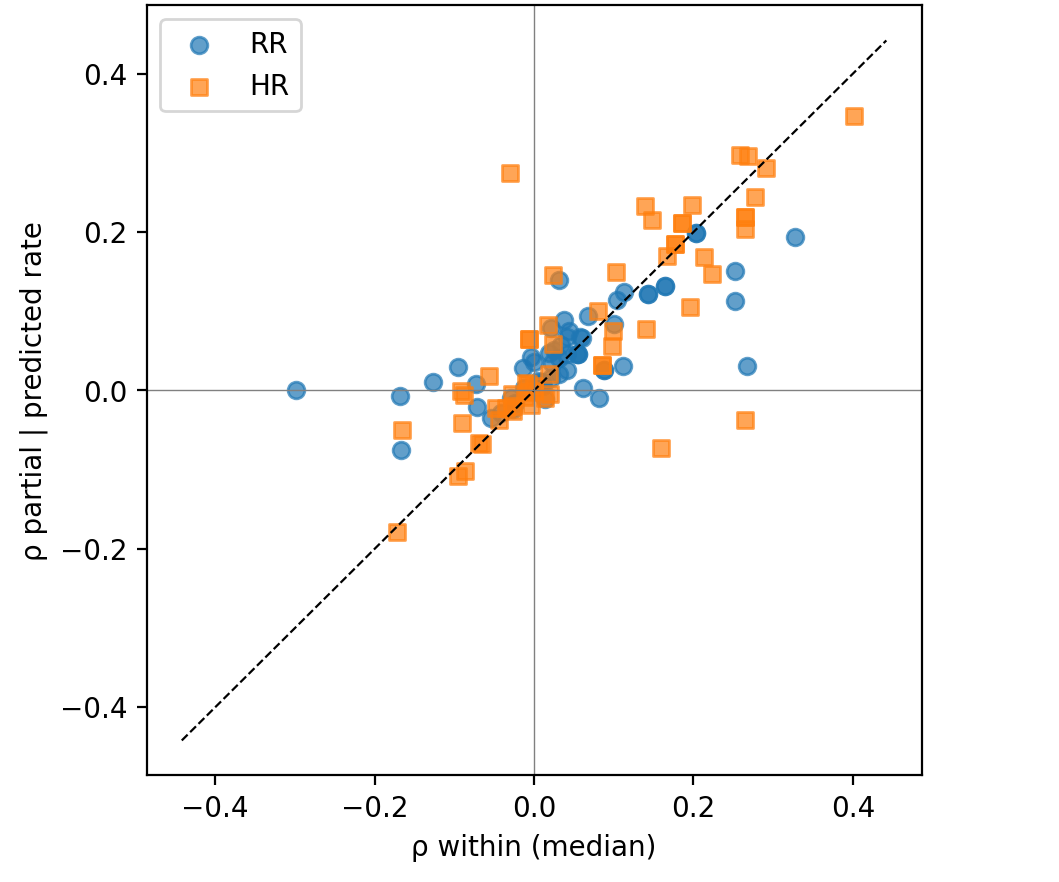}
    \caption{Median within-method $\rho$ (horizontal)
    against the median partial $\rho$ given the predicted
    rate (vertical), one point per indicator, task and
    dataset, including the Trainset. Points on the diagonal
    carry information beyond the rate; points pulled toward
    the horizontal axis carry it mainly through the rate.}
    \label{fig:partial}
\end{figure}

For RR, removing the worst 30\% by AMM lowered MAE by only 1.0--4.6\% relative to random removal. On MAHNOB, where the RR indicators correlated most strongly with error, the correlation largely disappeared once the predicted rate was controlled for: $Q_{\mathrm{HJ\text{-}C}}$ had a within-method $\rho$ of 0.27 but a partial $\rho$ of 0.03, and $Q_{\mathrm{HJ\text{-}M}}$ $-0.30$ and 0.00 (Fig.~\ref{fig:partial}). On COHFACE part of the signal remained ($Q_{\mathrm{HJ\text{-}M}}$ 0.33 to 0.19), but these indicators were dropped on the Trainset because their sign was not consistent across subjects, and filtering on COHFACE beat random removal by only 2.2\%. These indicators describe the frequency content of the waveform, which largely restates the rate the method already estimated. Significance tests make the split between datasets explicit (Supplement S6). On MAHNOB, raw within-method correlations were significant for $Q_{\mathrm{HJ\text{-}C}}$ (useful in 8 of 10 methods) and $Q_{\mathrm{HJ\text{-}M}}$ (harmful in 6), and none survived given the predicted rate. On COHFACE the opposite happened: given the rate, $Q_{\mathrm{SNR}}$, $Q_{\mathrm{TMCC}}$, $Q_{\mathrm{HFD}}$, $Q_{\mathrm{ZCR}}$ and $Q_{\mathrm{HJ\text{-}M}}$ were significant in the useful direction in 9--10 of 10 methods. These COHFACE effects are real but small (median partial $\rho$ 0.09--0.19): they ranked methods only weakly (cross-method $\rho \leq 0.14$), did not make filtering worthwhile (2.2\% better than random), and sharpened weighting with the Trainset pool increased COHFACE error by 8.2\% (Supplement S8), so they did not improve fusion.

\subsubsection{Across-method ranking}

\begin{figure*}[!t]
    \centering
    \includegraphics[width=\linewidth, keepaspectratio]{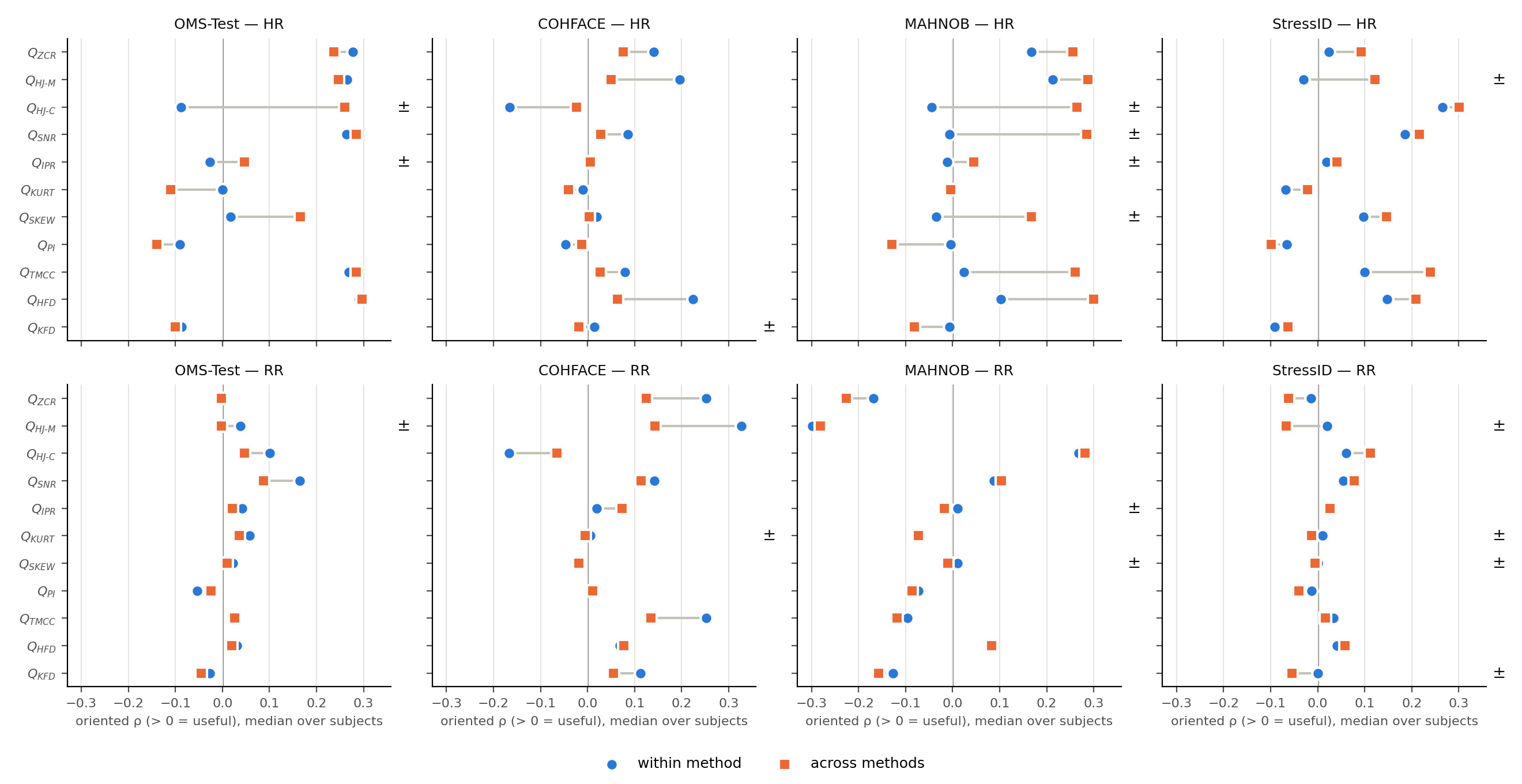}
    \caption{Spearman $\rho$ between indicator and absolute error, within methods (circles) and across methods in the same window (squares), median over subjects. Pool indicators are oriented by their Trainset direction so that $\rho>0$ means useful; dropped indicators are shown with the lower-is-better orientation. $\pm$ marks indicators whose within- and across-method $\rho$ differ in sign.}
    \label{fig:cross}
\end{figure*}

Fusion needs indicators to rank methods within a window. For HR they did on three of four test sets: cross-method $\rho$ was 0.26--0.30 for $Q_{\mathrm{SNR}}$, $Q_{\mathrm{TMCC}}$, $Q_{\mathrm{HJ\text{-}C}}$ and $Q_{\mathrm{HFD}}$ on OMS-Test and MAHNOB, and 0.21--0.30 for the same indicators on StressID (Fig.~\ref{fig:cross}). On COHFACE, cross-method $\rho$ was at most 0.08. For RR, the strongest consistent ranking signal was $Q_{\mathrm{SNR}}$ (cross-method $\rho$ 0.08--0.11); other indicators changed sign between datasets (e.g.\ $Q_{\mathrm{HJ\text{-}M}}$: 0.14 on COHFACE, $-0.28$ on MAHNOB).

\subsection{Quality-weighted and similarity-weighted fusion}\label{sec:res_fusion}\label{sec:res_weights}

Table~\ref{tab:fusion} gives MAE and $r$ for the ALL group, and Table~\ref{tab:boot} gives pooled bootstrap gains. The constant and oracle rows are discussed in Section~\ref{sec:controls}.

\textbf{HR.} Equal averaging was worse than the hindsight-best method ($-19.5$\% [$-33.4$, $-10.8$]) and no better than the method chosen in advance ($+4.5$\% [$-8.8$, 12.0]). Quality information improved on averaging: untuned AMM weighting by 4.9\% [2.9, 7.6], Trainset-tuned weighting by 10.9\% [5.2, 18.4], and the learned score L-Q by 9.2\% [4.1, 15.7]. Tuned weighting improved on Baseline-train by 15.6\% [8.1, 23.1] and was not distinguishable from the hindsight-best method ($-5.2$\% [$-12.2$, 2.0]); L-Q remained below it ($-8.0$\% [$-14.7$, $-2.1$]). The gain over Baseline-train comes mostly from COHFACE and MAHNOB, where the Trainset's best method (PF) transferred poorly (9.11 and 10.91\,bpm) while weighting stayed close to the best available method (7.32 and 7.22).

\textbf{RR.} Combining methods reduced error relative to single methods: the equal mean improved on the hindsight-best method by 5.6\% [1.9, 9.1] and on Baseline-train by 13.5\% [10.5, 16.8]. Quality weighting gave no further improvement. Untuned AMM and Trainset-SMM weighting stayed within 0.2\% of the equal mean, while sharper weighting increased error (tuned AMM $-2.6$\% [$-4.8$, $-0.8$]; L-Q+rate $-1.3$\% [$-2.5$, $-0.1$]). All fused RR estimates were essentially uncorrelated with the reference ($r$ between $-0.11$ and 0.22).

\textbf{Similarity.} Weighting methods by their agreement with the other methods gave the same result as averaging: soft similarity fusion was within 0.4\% of the equal mean for both tasks (HR $+0.4$\% [$-0.1$, 0.9]; RR $-0.2$\% [$-0.3$, $-0.0$]). Selecting the single most agreeing method was clearly worse than averaging (HR $-21.8$\% [$-31.0$, $-15.3$]; RR $-10.2$\% [$-12.7$, $-8.2$]), with the largest HR losses on MAHNOB ($-30$\%) and StressID ($-27$\%). Quality indicators, not agreement, account for the HR gains above.

\begin{table*}[!t]
\centering
\caption{Fusion and controls, ALL group: MAE with Pearson $r$ in parentheses. Mean is the unweighted mean over datasets. $^{\dagger}$Uses test labels (diagnostic only). $Q_{\mathrm{S}}$/$Q_{\mathrm{H}}$: quality-based soft/hard fusion; $S_{\mathrm{S}}$/$S_{\mathrm{H}}$: similarity-based soft/hard fusion; L-Q: learned score; $\tau^{*}$: Trainset-tuned temperature. $r$ is undefined for constant predictions.}
\label{tab:fusion}
\small
\begin{tabular}{lccccc}
\toprule
Row & OMS-Test & COHFACE & MAHNOB & StressID & Mean \\
\midrule\multicolumn{6}{l}{\textit{Heart rate (bpm)}}\\
Baseline (test-chosen)$^{\dagger}$ & 5.46 (0.70) & 7.29 (0.51) & 5.99 (0.62) & 13.07 (0.06) & 7.95 \\
Baseline-train & 6.01 (0.70) & 9.11 (0.41) & 10.91 (0.07) & 13.07 (0.06) & 9.78 \\
Equal mean & 6.75 (0.64) & 7.14 (0.55) & 8.66 (0.38) & 14.63 (0.07) & 9.29 \\
Median & 5.97 (0.67) & 6.87 (0.56) & 8.18 (0.39) & 14.12 (0.16) & 8.79 \\
Constant & 11.73 & 12.26 & 11.29 & 13.13 & 12.10 \\
\addlinespace[2pt]
$Q_{\mathrm{S}}$-AMM ($\tau=1$) & 6.20 (0.68) & 7.14 (0.57) & 8.03 (0.44) & 14.00 (0.08) & 8.84 \\
$Q_{\mathrm{S}}$-AMM ($\tau^{*}$) & 5.29 (0.72) & 7.32 (0.55) & 7.22 (0.52) & 13.46 (0.07) & 8.32 \\
Trainset-$Q_{\mathrm{S}}$-SMM & 6.32 (0.67) & 7.23 (0.56) & 8.24 (0.42) & 13.82 (0.08) & 8.90 \\
$Q_{\mathrm{H}}$-AMM & 5.46 (0.68) & 7.38 (0.49) & 7.44 (0.44) & 14.23 (0.01) & 8.63 \\
\addlinespace[2pt]
L-Q & 5.32 (0.72) & 7.18 (0.54) & 8.02 (0.44) & 13.34 (0.09) & 8.46 \\
L-Q+rate & 5.27 (0.73) & 7.19 (0.54) & 8.77 (0.35) & 13.84 (0.03) & 8.77 \\
$S_{\mathrm{S}}$-AMM & 6.70 (0.64) & 7.11 (0.55) & 8.62 (0.38) & 14.63 (0.08) & 9.27 \\
$S_{\mathrm{H}}$-AMM & 8.11 (0.48) & 7.87 (0.41) & 11.25 (0.26) & 18.58 (0.03) & 11.45 \\
\addlinespace[2pt]
Oracle$^{\dagger}$ & 2.54 (0.92) & 2.51 (0.93) & 2.59 (0.94) & 6.06 (0.77) & 3.42 \\
Null oracle$^{\dagger}$ & 3.60 & 3.30 & 3.44 & 6.22 & 4.14 \\
\midrule\multicolumn{6}{l}{\textit{Respiratory rate (brpm)}}\\
Baseline (test-chosen)$^{\dagger}$ & 4.55 (0.13) & 3.77 (0.03) & 5.24 (0.02) & 5.27 (0.06) & 4.71 \\
Baseline-train & 4.64 (0.16) & 4.27 ($-$0.07) & 6.01 (0.06) & 5.69 (0.07) & 5.15 \\
Equal mean & 4.33 (0.22) & 3.03 ($-$0.09) & 5.65 ($-$0.02) & 4.99 (0.13) & 4.50 \\
Median & 4.30 (0.22) & 3.19 ($-$0.11) & 5.91 ($-$0.03) & 5.14 (0.11) & 4.64 \\
Constant & 4.52 & 2.83 & 5.20 & 4.93 & 4.37 \\
\addlinespace[2pt]
$Q_{\mathrm{S}}$-AMM ($\tau=1$) & 4.31 (0.22) & 3.03 ($-$0.09) & 5.63 ($-$0.02) & 4.98 (0.13) & 4.49 \\
$Q_{\mathrm{S}}$-AMM ($\tau^{*}$) & 4.39 (0.15) & 3.27 ($-$0.07) & 5.58 ($-$0.03) & 5.08 (0.08) & 4.58 \\
Trainset-$Q_{\mathrm{S}}$-SMM & 4.31 (0.22) & 3.04 ($-$0.09) & 5.61 ($-$0.02) & 4.98 (0.12) & 4.48 \\
$Q_{\mathrm{H}}$-AMM & 4.68 (0.10) & 3.86 ($-$0.05) & 5.67 (0.01) & 5.42 (0.05) & 4.91 \\
\addlinespace[2pt]
L-Q & 4.32 (0.18) & 3.19 ($-$0.09) & 5.48 ($-$0.02) & 5.04 (0.06) & 4.51 \\
L-Q+rate & 4.31 (0.20) & 3.01 ($-$0.10) & 5.85 ($-$0.03) & 5.12 (0.08) & 4.57 \\
$S_{\mathrm{S}}$-AMM & 4.32 (0.22) & 3.03 ($-$0.10) & 5.67 ($-$0.02) & 5.01 (0.12) & 4.51 \\
$S_{\mathrm{H}}$-AMM & 4.43 (0.19) & 3.67 ($-$0.09) & 6.12 (0.00) & 5.44 (0.07) & 4.91 \\
\addlinespace[2pt]
Oracle$^{\dagger}$ & 1.27 (0.93) & 0.90 (0.89) & 1.76 (0.87) & 1.47 (0.91) & 1.35 \\
Null oracle$^{\dagger}$ & 1.33 & 0.76 & 1.64 & 1.40 & 1.28 \\
\bottomrule
\end{tabular}
\end{table*}

\begin{table}[!t]
\centering
\caption{Pooled bootstrap gains (75 subjects, four datasets, equal dataset weight). Positive values favour the second configuration. 95\% percentile confidence intervals in brackets.}
\label{tab:boot}
\small
\resizebox{\linewidth}{!}{%
\begin{tabular}{lcc}
\toprule
Comparison & HR & RR \\
\midrule
Baseline $\rightarrow$ Equal mean & $-19.5$ [$-33.4$, $-10.8$] & 5.6 [1.9, 9.1] \\
Baseline-train $\rightarrow$ Equal mean & 4.5 [$-8.8$, 12.0] & 13.5 [10.5, 16.8] \\
Constant $\rightarrow$ Equal mean & 24.0 [14.7, 31.4] & $-3.1$ [$-6.1$, $-0.4$] \\
\midrule
Equal mean $\rightarrow$ $Q_{\mathrm{S}}$-AMM ($\tau=1$) & 4.9 [2.9, 7.6] & 0.2 [$-0.1$, 0.4] \\
Equal mean $\rightarrow$ $Q_{\mathrm{S}}$-AMM ($\tau^{*}$) & 10.9 [5.2, 18.4] & $-2.6$ [$-4.8$, $-0.8$] \\
Equal mean $\rightarrow$ Trainset-$Q_{\mathrm{S}}$-SMM & 3.9 [1.7, 6.7] & 0.2 [$-0.1$, 0.5] \\
Equal mean $\rightarrow$ L-Q & 9.2 [4.1, 15.7] & $-0.8$ [$-2.9$, 0.9] \\
Equal mean $\rightarrow$ L-Q+rate & 6.3 [0.3, 13.3] & $-1.3$ [$-2.5$, $-0.1$] \\
Equal mean $\rightarrow$ $S_{\mathrm{S}}$-AMM & 0.4 [$-0.1$, 0.9] & $-0.2$ [$-0.3$, $-0.0$] \\
Equal mean $\rightarrow$ $S_{\mathrm{H}}$-AMM & $-21.8$ [$-31.0$, $-15.3$] & $-10.2$ [$-12.7$, $-8.2$] \\
\midrule
Baseline-train $\rightarrow$ $Q_{\mathrm{S}}$-AMM ($\tau^{*}$) & 15.6 [8.1, 23.1] & 11.6 [8.8, 14.6] \\
Baseline $\rightarrow$ $Q_{\mathrm{S}}$-AMM ($\tau^{*}$) & $-5.2$ [$-12.2$, 2.0] & 3.5 [$-0.1$, 6.9] \\
Baseline-train $\rightarrow$ L-Q & 14.3 [6.8, 20.9] & 13.1 [10.2, 16.0] \\
Baseline $\rightarrow$ L-Q & $-8.0$ [$-14.7$, $-2.1$] & 5.1 [1.6, 8.5] \\
\bottomrule
\end{tabular}}
\end{table}

\textbf{Weight concentration.} At $\tau=1$, weights were nearly uniform: the effective number of methods was 6.25--6.33 of 7 for HR and 9.67--9.70 of 10 for RR, which is why untuned AMM rows sit close to the equal mean. The Trainset-tuned temperatures concentrate the weights (about 3.0 effective HR methods and 3.4--3.5 RR methods), which helps HR, where the indicators rank methods, and hurts RR, where they do not. The Trainset curves were consistent with the HR test results but not with the RR ones (Supplement S4): on the Trainset, $\tau^{*}$ lowered HR MAE by 11.2\% (6.18 to 5.49\,bpm) and RR MAE by 4.1\% (3.83 to 3.67\,brpm) relative to the equal mean. The HR gain replicated on the test sets (10.9\%), whereas the RR gain reversed ($-2.6$\%), so the RR temperature was fitted to the five Trainset subjects rather than to transferable quality information. The learned models agree: for HR the largest coefficients were on $Q_{\mathrm{HFD}}$ and $Q_{\mathrm{TMCC}}$, the two indicators with the strongest within-method correlations on the Trainset; three smaller HR coefficients ($Q_{\mathrm{HJ\text{-}M}}$, $Q_{\mathrm{HJ\text{-}C}}$, $Q_{\mathrm{SNR}}$) took the opposite sign to their directions, consistent with collinearity among the Hjorth and fractal indicators. For RR, $Q_{\mathrm{SNR}}$ carried most of the weight, and neither learned score, with or without the predicted rate, improved on the equal mean.

\section{Fusion gains against controls}\label{sec:reassess}

The comparisons above are common in fusion studies, but they do not show whether a gain comes from quality information or from the three simpler mechanisms described in Section~\ref{sec:intro}: variance reduction by averaging, shrinkage toward the typical rate, and the chance advantage of an oracle that chooses among many estimates. This section tests the results of Section~\ref{sec:results} against these mechanisms.

\subsection{Averaging, typical-rate and oracle controls}\label{sec:controls}

\textbf{Controls.} \emph{Constant} predicts the Trainset median reference rate (78.95\,bpm; 12.52\,brpm) in every window. The \emph{Oracle} picks, per window, the method with the smallest error. The \emph{Null oracle}, a permutation-based control, recomputes it after permuting each method's estimates across all windows of the test set, independently for each method and leaving the reference in place, and reports the mean over 200 permutations. It is the error that picking the best of the methods per window reaches by chance, so the difference between Oracle and Null oracle is the part of the oracle's advantage that comes from each estimate belonging to its own window, rather than from choosing the closest of many estimates by chance. This includes each method's own tracking of the reference as well as any complementarity between methods. Oracle rows use test labels and are diagnostic only.

\textbf{HR.} The HR fusion results remain above these controls. The equal mean improved on the constant predictor by 24.0\% [14.7, 31.4], and quality weighting improved further on the equal mean (Section~\ref{sec:res_fusion}). The oracle was 0.8--1.1\,bpm better than its null on OMS-Test, COHFACE and MAHNOB (e.g.\ 2.54 vs.\ 3.60\,bpm on OMS-Test) and 0.16\,bpm on StressID, so the HR estimates carry window-specific information that a selector can exploit.

\begin{figure*}[!t]
    \centering
    \includegraphics[width=\linewidth, keepaspectratio]{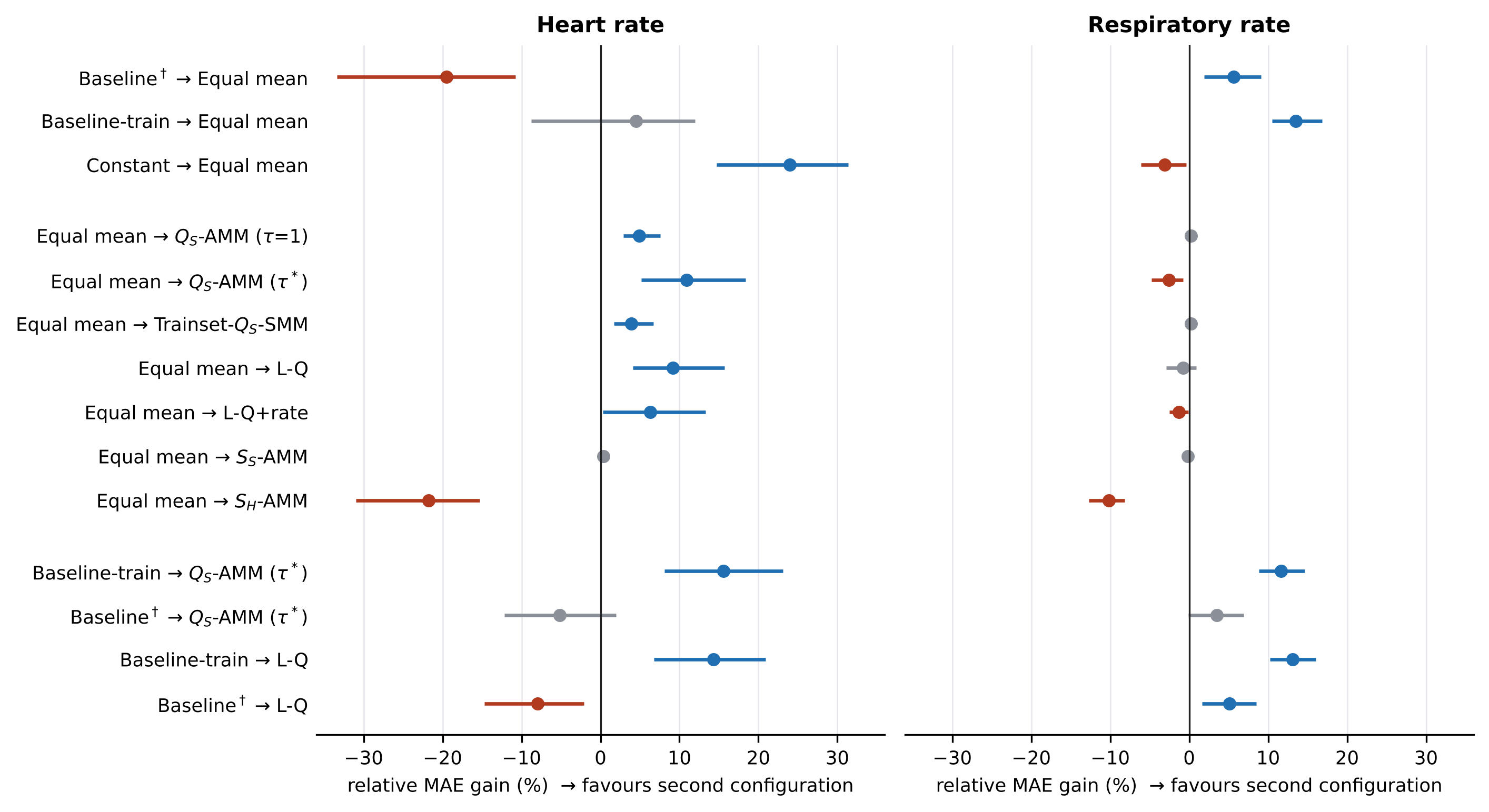}
    \caption{Pooled bootstrap gains (Table~\ref{tab:boot}) with 95\% confidence intervals; right of zero favours the second configuration of each pair (A $\rightarrow$ B). Coloured intervals exclude zero. $^{\dagger}$Chosen with test labels.}
    \label{fig:forest}
\end{figure*}

\textbf{RR.} The constant predictor was more accurate than the equal mean ($-3.1$\% [$-6.1$, $-0.4$]); on COHFACE it was more accurate than every method and every fusion rule (2.83\,brpm). The improvement from averaging therefore reflects shrinkage toward the typical rate rather than better tracking of breathing: averaging several uncorrelated estimates pulls the result toward the centre of the band, where most true rates lie. The oracle was never more than 0.06\,brpm better than its null and was worse on three datasets (e.g.\ COHFACE 0.90 vs.\ 0.76\,brpm): the null oracle gives no evidence that the RR estimates carry window-specific information, and the low RR oracle error reflects picking the closest of ten estimates by chance. Figure~\ref{fig:forest} summarises these comparisons.

\subsection{Sensitivity to the respiratory reference}\label{sec:res_robust}

The full RR pipeline was repeated with peak intervals on the belt as the reference. Every Trainset choice was re-derived under this reference with the same procedure: directions and pool, SMM subsets, ridge models, temperature and the constant (Supplement S9). HR was not affected. The two belt references agreed only moderately (spectral vs.\ peak: $r=0.35$--0.51, MAE 2.7--4.5\,brpm), so reference uncertainty in 10-s windows is of the same order as method error. The peak reference also reads higher (Trainset median 14.1 vs.\ 12.5\,brpm). Under it, the Trainset kept eight RR indicators instead of four, adding $Q_{\mathrm{ZCR}}$, $Q_{\mathrm{HJ\text{-}M}}$, $Q_{\mathrm{HJ\text{-}C}}$, $Q_{\mathrm{PerI}}$ and $Q_{\mathrm{KFD}}$ and dropping $Q_{\mathrm{KURT}}$; the Hjorth, zero-crossing and Katz indicators had cross-method $\rho$ of 0.21--0.28 on the Trainset.

Quality weighting then lowered RR error relative to averaging (L-Q+rate 11.2\% [5.0, 16.0]; Trainset-SMM 4.5\% [2.4, 6.2]), but three observations show that the gain is not quality information. First, every fused estimate remained uncorrelated with the reference ($r \leq 0.17$), and none outperformed the constant predictor on average over the four datasets (equal mean vs.\ constant $-21.8$\% [$-27.3$, $-16.7$]). Second, the weighted estimates moved toward the constant rather than toward the reference (e.g.\ MAHNOB: tuned AMM 4.36 vs.\ constant 4.31\,brpm), and the newly kept indicators had within-method correlations close to zero once the predicted rate was controlled for (Trainset partial $\rho$ $-0.13$ to 0.09; $-0.04$ to 0.02 on MAHNOB and StressID). Weighting that favours methods whose estimates lie near the reference's typical level produces exactly this pattern. Third, on COHFACE, whose rate-series reference is the same in both analyses, these indicators pointed the wrong way (cross-method $\rho$ $-0.14$ to $-0.01$) and the same weighting increased error (tuned AMM 4.29 vs.\ 3.03\,brpm for the equal mean).

\subsection{Sensitivity to window length}\label{sec:res_window}

Because a 10-s window holds only 1--3 breaths, the RR analysis was repeated with 30-s windows (1-s step); HR was not changed. The RR reference becomes FFT on the belt over the 30-s window (COHFACE: the mean of the rate series over the window), and every Trainset choice was re-derived with the same procedure (Supplement S10).

With 30-s windows, single RR methods and the reference improved. The Trainset chose Welch instead of MUSIC (mean MAE 3.77 vs.\ 3.88\,brpm). The best single method per test set averaged 4.18\,brpm (4.71 at 10\,s), and correlations with the reference reached $r=0.28$ (Mot-F, OMS-Test) and 0.21 (CP, StressID). The two belt references agreed far better ($r=0.70$--0.82, MAE 1.5--3.1\,brpm, vs.\ 0.35--0.51 and 2.7--4.5\,brpm at 10\,s), and $Q_{\mathrm{SNR}}$ correlated more strongly with a method's own error ($|\rho|$ 0.27 vs.\ 0.16 on OMS-Test; 0.23 vs.\ 0.09 on MAHNOB).

Fusion did not follow. Under the Trainset rules, $Q_{\mathrm{SNR}}$ was dropped because one of the five Trainset subjects had the opposite sign, leaving $Q_{\mathrm{KURT}}$, $Q_{\mathrm{SKEW}}$ and $Q_{\mathrm{PerI}}$. Untuned weighting matched the equal mean ($-0.0$\% [$-0.5$, 0.4]), tuned weighting increased error ($-6.9$\% [$-11.5$, $-2.9$]), the oracle was no better than its null on three of four datasets (mean 1.24 vs.\ 1.26\,brpm), and the constant predictor still matched averaging ($-3.3$\% [$-8.5$, 1.6]), most clearly on MAHNOB (4.92 vs.\ 5.75\,brpm).

Because this removed $Q_{\mathrm{SNR}}$ from the pool, the 30-s analysis was repeated with the indicator pool and directions fixed at 10\,s ($Q_{\mathrm{SNR}}$, $Q_{\mathrm{KURT}}$, $Q_{\mathrm{SKEW}}$, $Q_{\mathrm{HFD}}$), re-tuning only $\tau^{*}$ and the ridge model on the 30-s Trainset. This did not change the result. No variant beat the equal mean (untuned $+0.1$\% [$-0.2$, 0.4]; tuned $-2.2$\% [$-5.0$, 0.4]; L-Q $+1.0$\% [$-0.8$, 2.7]) or the constant predictor. $Q_{\mathrm{SNR}}$ ranked methods in the expected direction on every test set, but weakly (cross-method $\rho$ 0.08--0.16, against 0.21--0.30 for the HR indicators), and neither weighting by $Q_{\mathrm{SNR}}$ alone ($+0.6$\% [$-1.0$, 2.1]) nor a ridge model on all 11 indicators ($+1.9$\% [$-1.0$, 4.4]), both run as exploratory checks, beat averaging. Of 24 group-level comparisons with the equal mean, one exploratory comparison favoured fusion (Supplement S10).

\section{Indicator usefulness for HR and RR}\label{sec:why}

A 10-s window holds 10--15 heartbeats but only 1--3 breaths. With many cycles, waveform regularity ($Q_{\mathrm{TMCC}}$), spectral concentration ($Q_{\mathrm{SNR}}$) and complexity ($Q_{\mathrm{HFD}}$) describe whether a clean periodic signal is present, and a clean periodic signal usually gives the right rate. With one or two cycles, the same measures may mainly describe the frequency content of the window, which is the rate the method already estimated.

The partial correlations make this concrete: the strongest RR indicator effects (MAHNOB) vanished once the predicted rate was controlled for, whereas most HR effects persisted or grew. Where RR indicators did carry information beyond the rate (COHFACE), it was too weak to rank methods. The reference itself is also uncertain in short windows: the two belt references disagreed by 2.7--4.5\,brpm at 10\,s and by 1.5--3.1\,brpm at 30\,s.

Window length limits the RR estimates but is not the only constraint: with 30-s windows the estimates, the reference and $Q_{\mathrm{SNR}}$ all improved, yet no indicator ranked methods well enough for fusion to beat averaging. These results support an influence of cycle count but do not establish a sufficient window length or number of cycles for RR quality assessment. Indicators that assess breathing itself, such as breath-by-breath morphology~\cite{Romano2023Signal}, breath-based windows and more accurate references may provide information that the evaluated indicators miss.

\section{Discussion}\label{sec:discussion}

\textbf{Quality weighting is useful for HR.} HR indicators identified a method's inaccurate windows (filtering 12--25\% better than random) and ranked methods within a window (cross-method $\rho$ up to 0.30), and the per-window oracle beat its null. With weights fixed on five subjects, tuned weighting reduced HR error by 11\% relative to averaging and by 16\% relative to the method a practitioner would have chosen in advance. The practical value is that the result depends less on that choice: the Trainset's best method failed to transfer to COHFACE and MAHNOB, while quality weighting stayed close to whichever method was best on each dataset. Fusion did not beat the best method chosen in hindsight, and should not be expected to when one method clearly dominates.

\textbf{For RR, apparent fusion gains reflect shrinkage.} Averaging RR estimates reduced error relative to single methods by 6--14\%, but the constant predictor outperformed averaging, the oracle was no better than its null, every fused estimate had $r \leq 0.22$, and sharpening the weights made results worse. Under a second reference, weighting appeared to help, but only by moving estimates toward that reference's typical level; the same weights hurt on a dataset with an independent reference. A practitioner who reads an RR fusion gain as evidence of quality information would be misled; comparing with a constant predictor and a null oracle exposes this.

\textbf{Agreement and averaging.} A method's mean similarity to the others measures its distance from the consensus, and the equal mean already estimates the consensus; soft agreement weighting therefore returns close to the average. Selecting only the most agreeing method keeps the consensus but loses the error reduction that averaging gives. When the methods share an error, agreement also rewards that error: all HR methods compressed estimates toward 70--80\,bpm (Fig.~\ref{fig:hrdep}), so the method closest to the others is not the one closest to the truth. This is consistent with hard selection losing most on MAHNOB and StressID, but it remains an interpretation. Consensus is not useless: as a pseudo-reference it recovered all five HR indicator directions without labels (Section~\ref{sec:res_train}). Agreement can therefore help decide which indicators to trust, but weighting methods by agreement gives little more than averaging.

\textbf{Methodological lessons.} Two controls changed our conclusions during this work and should be standard. A common reference computed independently of the methods' estimator: an estimator-matched reference lowered RR MAE by up to 0.7\,brpm for MUSIC and 1.2\,brpm for Peak (Supplement S2.3). A constant predictor and a null oracle: without them, the RR results would have read as successful fusion with complementary methods.

\section{Limitations}\label{sec:limitations}

The Trainset, OMS-Test and StressID have five, six and five subjects; subject-level tests on these sets cannot reach significance after correction. Because pooled estimates weight the four datasets equally, StressID carries a quarter of the pooled weight; weighting every subject equally instead (a ratio of summed per-subject MAEs, in which COHFACE and MAHNOB hold about four-fifths of the weight) does not change which configuration is better in any comparison, but reduces the HR gain of tuned weighting over averaging from 10.9\% to 7.9\% [2.9, 13.5], because that gain is absent on COHFACE and largest on OMS-Test, whose share falls to 7\% (Supplement S8). HR was evaluated in 10-s windows only; RR was evaluated in 10-s and 30-s windows, but not with longer windows or breath-by-breath segmentation, and the 30-s analysis also changed the RR estimator and reference, so the effect of window length alone cannot be isolated. Filtering is offline over the full recording; causal, real-time filtering was not evaluated. Deep models were trained at 30\,fps on OMuSense-23 and run at that rate by resampling; models trained on each dataset's native rate might behave differently. All HR methods compress the rate range toward 70--80\,bpm, and fusion cannot correct an error that all candidates share. Finally, the reference rules were validated against longer-window rates from the same contact sensors rather than against manual annotation.

\section{Conclusion}\label{sec:conclusion}

Reference-free quality indicators help video-based HR fusion: with all choices fixed in advance, quality weighting reduced HR error by 11\% relative to averaging and by 16\% relative to the method chosen in advance, approaching the best method chosen in hindsight. For RR in 10-s windows, no indicator improved fusion; averaging helped only by shrinking toward the typical rate, a constant predictor did better, and apparent quality gains under a second reference reflected estimates moving toward that reference's typical rate rather than better signal quality. Lengthening RR windows to 30\,s improved the estimates and the reference but not quality-guided fusion, even with $Q_{\mathrm{SNR}}$ restored to the pool. Evaluations of quality-guided fusion should report equal averaging, a constant predictor and a null oracle alongside the best single method, with choices fixed before testing and a reference independent of the methods' estimator.

\printcredits

\bibliographystyle{cas-model2-names}
\bibliography{references}

\FloatBarrier
\clearpage
\onecolumn
\appendix
\noindent{\Large\bfseries Supplementary material}\par\bigskip
\suppressfloats[t]
\renewcommand{\topfraction}{0.95}
\renewcommand{\textfraction}{0.05}
\renewcommand{\floatpagefraction}{0.85}
\setcounter{topnumber}{3}
\setcounter{totalnumber}{4}
\setcounter{section}{0}
\renewcommand{\thesection}{S\arabic{section}}
\renewcommand{\thesubsection}{S\arabic{section}.\arabic{subsection}}
\counterwithin{table}{section}\renewcommand{\thetable}{S\arabic{section}.\arabic{table}}
\counterwithin{figure}{section}\renewcommand{\thefigure}{S\arabic{section}.\arabic{figure}}
\setcounter{table}{0}\setcounter{figure}{0}
\noindent This supplement follows the structure of the main text. Part~A gives additional methods and Part~B additional results. Each section starts with the main-text section it supports. Tables and figures are numbered by section (e.g.\ Table~S2.1 is the first table of Section~S2). The sections are listed below.

\begin{center}
\small
\textbf{Contents of the supplement}\par\smallskip
\begin{tabular}{@{}lll@{}}
\toprule
& Section & Main text \\
\midrule
\multicolumn{3}{@{}l}{\textit{A. Additional methods}} \\
S1 & Signal filtering and model frame rates & 3.1 \\
S2 & Reference rates: quality checks, validation, estimator-matched reference, MAHNOB-HCI ECG lead & 3.1 \\
S3 & Indicator directions, subsets and models fixed on the Trainset & 3.4, 4.2 \\
S4 & Fusion temperature and weight sharpness & 3.4, 4.3 \\
\multicolumn{3}{@{}l}{\textit{B. Additional results}} \\
S5 & Single-method results & 4.1 \\
S6 & Quality indicators and estimation error & 4.2 \\
S7 & Fusion results by method group & 4.3 \\
S8 & Fusion gains by dataset and dataset weighting & 4.3 \\
S9 & Alternative respiratory reference & 5.2 \\
S10 & RR estimation with 30-s windows & 5.3 \\
\bottomrule
\end{tabular}
\end{center}

\noindent\textbf{Notation.} Baseline = best single method on each test set (chosen in hindsight); Baseline-train = best on the Trainset; $Q$/$S$ = quality/similarity score; H/S = hard/soft fusion; AMM = all indicators of the pool; SMM = Trainset-selected subset; L-Q = learned ridge score; $\tau^{*}$ = Trainset-tuned temperature. $^{\dagger}$Uses test labels.

\bigskip\noindent{\large\bfseries A. Additional methods}\par\medskip

\section{Signal filtering and model frame rates}\label{sec:s1}
\textit{Supports main-text Section~3.1.} How the filter chains were chosen on the Trainset, and how the deep models are run.

\subsection{Filter chains}
Table~\ref{tab:s2filters} lists, for each method group, the chain selected on the Trainset grid, the best-scoring chain, and the rule that decided between them.

\begin{table}[!htp]
\centering
\caption{Filter chains selected on the Trainset grid.}
\label{tab:s2filters}
\footnotesize
\begin{adjustbox}{max width=\linewidth}
\begin{tabular}{p{2.6cm}p{4.2cm}p{4.2cm}rp{4.6cm}}
\toprule
Family & Selected chain & Best-scoring chain & Excess (pp) & Selection rule \\
\midrule
Classical rPPG & median spike removal $\rightarrow$ FIR band-pass (0.7--3.0\,Hz) & Savitzky--Golay pre-smoothing $\rightarrow$ FIR band-pass (0.7--3.0\,Hz) & 0.54 & tie-break: tied with SG pre-smoothing ($<$ 1 pp excess); median spike removal keeps the pulse waveform (morphology indicators) and does not attenuate high HR \\
Deep rPPG & spike removal $\rightarrow$ Butterworth high-pass (0.35\,Hz) $\rightarrow$ FIR band-pass & same as selected & 0.00 & best combined rank \\
Classical respiratory & Butterworth band-pass $\rightarrow$ Savitzky--Golay smoothing & Wiener pre-denoising $\rightarrow$ Butterworth band-pass $\rightarrow$ Savitzky--Golay smoothing & 0.71 & simplest within 1.0 pp of best \\
Deep and motion respiratory & second-order Butterworth band-pass (0.08--0.50\,Hz) & Wiener pre-denoising $\rightarrow$ Butterworth band-pass & 0.61 & simplest within 1.0 pp of best \\
\bottomrule
\end{tabular}\end{adjustbox}
\par\smallskip\begin{minipage}{\linewidth}\footnotesize\textit{Note.} Candidates were scored by mean MAE rank and mean relative MAE excess over the best candidate across methods, all four estimators, and full-recording and window-wise filtering. Scoring on full-recording filtering alone gives the same selection. For classical rPPG, Savitzky--Golay pre-smoothing scores best under both scores, by 0.54\,pp (combined) and 0.05\,pp (full recording) over median spike removal; both margins are below 1\,pp, so the tie-break keeps median spike removal. Excess = percentage points above the best-scoring chain.\end{minipage}
\end{table}

Selected chain parameters: median spike removal with a 0.15-s kernel; FIR band-pass 0.7--3.0\,Hz; second-order Butterworth high-pass at 0.35\,Hz (deep rPPG); second-order Butterworth band-pass 0.08--0.50\,Hz; Savitzky--Golay smoothing with a 1.5-s window and order 3 (classical respiratory). Reference signals: fourth-order Butterworth band-pass (HR 0.7--3.0\,Hz, RR 0.08--0.50\,Hz). All filters are applied to the full recording before windowing.

\subsection{Deep-model frame rate}
All deep models were trained at 30\,fps and are run at 30\,fps: MAHNOB-HCI (61\,fps) and StressID (62\,fps) frames are subsampled, COHFACE (20\,fps) frames are upsampled by linear blending of neighbouring frames, and model outputs are interpolated back to the native frame rate. Clips of each model's native length (180--300 frames) are processed with 50\% overlap, standardised and merged with a Hann cross-fade; recording ends are mirror-padded.

\section{Reference rates}\label{sec:s2}
\textit{Supports main-text Section~3.1.} Quality control of the common references (S2.1), their validation against longer-window rates (S2.2), how far an estimator-matched reference departs from them (S2.3), and the MAHNOB-HCI ECG lead correction (S2.4).

\subsection{Quality checks}
Tables~\ref{tab:s1qchr} and~\ref{tab:s1qcrr} report how many reference windows each quality check flagged. No HR window failed on any dataset. For RR, 2.5--9.1\% of belt windows had their spectral peak at a band edge.

\begin{table}[!htp]
\centering
\caption{HR reference quality control per dataset. Beat rule on contact PPG; R-R rule on ECG.}
\label{tab:s1qchr}
\footnotesize
\begin{adjustbox}{max width=\linewidth}
\begin{tabular}{llrrrrrrrr}
\toprule
Dataset & Rule & Windows & \% fail & \% peaks & \% fallback & \% $\geq$50\% covered & \% R-R rejected & Notched rec. & Median intervals \\
\midrule
Trainset & beats (PPG) & 3117 & 0.0 & 100.00 & 0.00 & -- & -- & -- & 12 \\
OMS-Test & beats (PPG) & 3696 & 0.0 & 99.95 & 0.05 & -- & -- & -- & 10 \\
COHFACE & beats (PPG) & 7680 & 0.0 & 100.00 & 0.00 & -- & -- & -- & 10 \\
MAHNOB & R-R (ECG) & 1968 & 0.0 & -- & -- & 100.0 & 0.16 & 0 & 13 \\
StressID & R-R (ECG) & 2940 & 0.0 & -- & -- & 100.0 & 0.52 & 1 & 14 \\
\bottomrule
\end{tabular}\end{adjustbox}
\end{table}

\begin{table}[!htp]
\centering
\caption{RR reference quality control per dataset.}
\label{tab:s1qcrr}
\footnotesize
\begin{adjustbox}{max width=\linewidth}
\begin{tabular}{llrrrrrrrr}
\toprule
Dataset & Main & Windows & \% band edge & \% weak peak & Peak/median & \% peaks & \% fallback & \% crossing & Median intervals \\
\midrule
Trainset & FFT & 3117 & 8.1 & 10.3 & 9.2 & 66.5 & 20.0 & 13.5 & 1 \\
OMS-Test & FFT & 3696 & 9.1 & 14.8 & 7.7 & 70.7 & 22.3 & 7.1 & 1 \\
COHFACE & rate series & 7680 & -- & -- & -- & -- & -- & -- & -- \\
MAHNOB & FFT & 1968 & 4.0 & 14.6 & 6.9 & 90.3 & 9.0 & 0.7 & 2 \\
StressID & FFT & 2940 & 2.5 & 27.2 & 4.5 & 81.5 & 16.6 & 1.8 & 2 \\
\bottomrule
\end{tabular}\end{adjustbox}
\par\smallskip\begin{minipage}{\linewidth}\footnotesize\textit{Note.} Left: Band edge, weak peak and Peak/median refer to FFT on the belt (main reference): windows whose spectral peak lies at a band edge, windows whose peak is below three times the median in-band power, and the median ratio of peak to median in-band power. The remaining columns refer to peak intervals on the belt (sensitivity reference): windows with a valid interval, with the fallback peak rule, with the timed zero-crossing rate, and the median number of breath intervals per 10-s window. COHFACE uses the provided rate series.\end{minipage}
\end{table}
\subsection{Validation against longer-window rates}
Each 10-s reference was compared with a rate from a longer, independent window on the same contact signal. For HR, the beat/R-R rule agreed best with the 30-s rate on every dataset (Table~\ref{tab:s1hrcheck}). For RR, FFT on the belt had a lower MAE than peak intervals on every dataset, so FFT was kept as the main reference (Table~\ref{tab:s1rrcheck}).

\begin{table}[!htp]
\centering
\caption{HR reference check against an independent 30-s rate.}
\label{tab:s1hrcheck}
\footnotesize
\begin{adjustbox}{max width=\linewidth}
\begin{tabular}{llrcccc}
\toprule
Dataset & Rule & Windows & Beat/R-R & FFT & Welch & MUSIC \\
\midrule
Trainset & beats (PPG) & 2725 & 1.45 (85.9) & 5.21 (67.4) & 5.54 (62.5) & 4.61 (72.0) \\
OMS-Test & beats (PPG) & 2958 & 1.19 (89.7) & 3.05 (75.5) & 3.39 (71.9) & 2.56 (80.3) \\
COHFACE & beats (PPG) & 4450 & 0.41 (98.5) & 0.98 (92.2) & 1.43 (89.1) & 0.65 (95.8) \\
MAHNOB & R-R (ECG) & 1008 & 1.48 (86.9) & 8.04 (78.6) & 9.04 (73.3) & 8.16 (79.0) \\
StressID & R-R (ECG) & 1740 & 2.48 (73.2) & 15.92 (54.3) & 18.74 (45.9) & 15.66 (50.2) \\
\midrule All & & 12881 & 1.17 (89.5) & 4.92 (77.0) & 5.68 (72.4) & 4.54 (79.7) \\
\bottomrule
\end{tabular}\end{adjustbox}
\par\smallskip\begin{minipage}{\linewidth}\footnotesize\textit{Note.} MAE (bpm) of each 10-s candidate against an independent 30-s comparison rate on the same contact signal; percentage of windows within 3\,bpm in parentheses. The beat/R-R rule is the common reference.\end{minipage}
\end{table}

\begin{table}[!htp]
\centering
\caption{RR reference check against breath-local 30-s rates.}
\label{tab:s1rrcheck}
\footnotesize
\begin{adjustbox}{max width=\linewidth}
\begin{tabular}{lrrrrrrr}
\toprule
Dataset & Windows & \% agreed & Peak & FFT & Welch & MUSIC & Crossing \\
\midrule
Trainset & 2817 & 68.9 & 2.96 & 1.30 & 1.57 & 1.28 & 1.44 \\
OMS-Test & 3336 & 58.8 & 3.76 & 1.58 & 1.81 & 1.51 & 1.74 \\
MAHNOB & 1008 & 66.4 & 2.32 & 1.58 & 1.88 & 1.66 & 1.50 \\
StressID & 1740 & 40.5 & 3.86 & 2.01 & 2.25 & 1.85 & 2.14 \\
\bottomrule
\end{tabular}\end{adjustbox}
\par\smallskip\begin{minipage}{\linewidth}\footnotesize\textit{Note.} MAE (brpm) of each 10-s belt estimate against breath-local rates from centred 30-s windows; \% agreed = windows where the comparison rates agree with each other. Decision rule fixed before the check: Peak minus FFT MAE = 1.79 [1.54, 2.03]\,brpm over the 141 belt recordings of 41 subjects (Trainset 15, OMS-Test 18, MAHNOB 48, StressID 60; bootstrap over recordings, which treats recordings of the same subject as independent, so the interval is somewhat narrow), so FFT was chosen. MUSIC was excluded because the methods use MUSIC.\end{minipage}
\end{table}
\subsection{Estimator-matched and common references}
This check asks how much the results depend on the estimator used to compute the reference. Table~\ref{tab:s1sens} compares each estimator, applied to the reference signal, with the common reference. Table~\ref{tab:s1matched} shows how single-method MAE changes when each estimator's predictions are scored against a reference from the same estimator. On the PPG datasets the HR changes stay within 0.6\,bpm.

\begin{table}[!htp]
\centering
\caption{Rate estimators applied to the reference signal, against the common reference.}
\label{tab:s1sens}
\footnotesize
\begin{adjustbox}{max width=\linewidth}
\begin{tabular}{lccccc}
\toprule
Estimator & Trainset & OMS-Test & COHFACE & MAHNOB & StressID \\
\midrule
\multicolumn{6}{l}{\textit{HR}} \\
FFT & 5.03 (0.76) & 3.54 (0.86) & 0.89 (0.99) & -- & -- \\
Welch & 5.41 (0.76) & 4.02 (0.84) & 1.37 (0.98) & -- & -- \\
MUSIC & 4.44 (0.79) & 3.02 (0.88) & 0.64 (0.99) & -- & -- \\
Peak (common reference) & 0.00 (1.00) & 0.00 (1.00) & 0.00 (1.00) & -- & -- \\
FFT on band-passed ECG & -- & -- & -- & 7.38 (0.55) & 14.38 (0.19) \\
Welch on band-passed ECG & -- & -- & -- & 8.94 (0.54) & 17.35 (0.12) \\
MUSIC on band-passed ECG & -- & -- & -- & 7.43 (0.54) & 13.86 (0.18) \\
Peak on band-passed ECG & -- & -- & -- & 4.82 (0.74) & 9.59 (0.47) \\
R-R (common reference) & -- & -- & -- & 0.00 (1.00) & 0.00 (1.00) \\
\midrule
\multicolumn{6}{l}{\textit{RR}} \\
FFT (common reference) & 0.00 (1.00) & 0.00 (1.00) & -- & 0.00 (1.00) & 0.00 (1.00) \\
Welch & 1.11 (0.90) & 1.42 (0.86) & -- & 1.51 (0.80) & 1.90 (0.80) \\
MUSIC & 1.01 (0.95) & 1.21 (0.94) & -- & 1.38 (0.91) & 1.69 (0.90) \\
Peak & 3.01 (0.48) & 4.04 (0.38) & -- & 2.84 (0.42) & 4.41 (0.35) \\
\bottomrule
\end{tabular}\end{adjustbox}
\par\smallskip\begin{minipage}{\linewidth}\footnotesize\textit{Note.} MAE with Pearson $r$ in parentheses. On ECG (MAHNOB, StressID) the spectral estimators and a generic peak rule are applied to the band-passed ECG. Rows marked common reference are zero by construction. COHFACE RR uses the provided rate series and has no belt-derived alternative.\end{minipage}
\end{table}

\begin{table}[!htp]
\centering
\caption{Change in single-method MAE with an estimator-matched reference.}
\label{tab:s1matched}
\footnotesize
\begin{adjustbox}{max width=\linewidth}
\begin{tabular}{llrrrrr}
\toprule
Task & Estimator & Trainset & OMS-Test & COHFACE & MAHNOB & StressID \\
\midrule
\multirow{4}{*}{HR} & FFT & 0.32 & $-$0.26 & 0.14 & 5.75 & 6.48 \\
 & Welch & $-$0.01 & $-$0.31 & 0.02 & 6.48 & 8.16 \\
 & MUSIC & 0.53 & $-$0.15 & 0.12 & 6.02 & 6.30 \\
 & Peak & 0.00 & 0.00 & 0.00 & 2.29 & $-$0.59 \\
\midrule
\multirow{4}{*}{RR} & FFT & 0.00 & 0.00 & 0.00 & 0.00 & 0.00 \\
 & Welch & 0.01 & 0.03 & 0.00 & 0.02 & 0.07 \\
 & MUSIC & $-$0.44 & $-$0.59 & 0.00 & $-$0.71 & $-$0.61 \\
 & Peak & $-$0.63 & $-$1.22 & 0.00 & $-$0.87 & $-$0.86 \\
\bottomrule
\end{tabular}\end{adjustbox}
\par\smallskip\begin{minipage}{\linewidth}\footnotesize\textit{Note.} Estimator-matched minus common reference: change in mean single-method MAE when each estimator's predictions are scored against a reference computed with the same estimator. Negative values mean the matched reference makes methods look better. The row equal to the common rule is zero by construction. On ECG (MAHNOB-HCI, StressID) the spectral estimators are poor HR references (Table~\ref{tab:s1sens}), so a matched reference there makes methods look worse rather than better; for RR the matched reference flatters MUSIC and Peak on every belt dataset.\end{minipage}
\end{table}
\subsection{MAHNOB-HCI ECG lead}
The ground-truth ECG distributed with our MAHNOB-HCI subset was lead EXG1. Its median QRS prominence was below 5 in 16 of 49 recordings (sessions 556, 560, 688, 1042, 1698, 1700, 2114, 2120, 2736, 2740, 3136, 3152, 3514, 3522, 3778, 3794), where R peaks could not be detected reliably. The 49 recordings include the session later excluded because its video is shorter than the reference (48 are analysed). For every recording, the lead among EXG1--EXG3 with the highest median QRS prominence was used instead (EXG3 in 45 recordings, EXG2 in 4), cropped exactly as the original signal (30-s pre-stimulus offset, verified by correlation with the stored signal). In one StressID recording, narrowband interference near 6\,Hz masked the QRS complex; spectral lines were notched before R-peak detection, applied only when no QRS could be detected otherwise.

\section{Indicator directions, subsets and models}\label{sec:s3}
\textit{Supports main-text Sections~3.4 and~4.2.} The indicator directions, subsets, temperatures and ridge models fixed on the Trainset before any test set was evaluated.

\begin{table}[!htp]
\centering
\caption{Indicator directions inferred on the Trainset.}
\label{tab:s5dir}
\footnotesize
\begin{tabular}{lcrcc}
\toprule
Indicator & Dir. & $\rho_{\times}$ & Subject signs & Label-free \\
\midrule
\multicolumn{5}{l}{\textit{HR}} \\
$Q_{\mathrm{ZCR}}$ & -- & 0.11 & + + $-$ + + & $\downarrow$ \\
$Q_{\mathrm{HJ\text{-}M}}$ & $\downarrow$ & 0.15 & + + + + + & $\downarrow$ \\
$Q_{\mathrm{HJ\text{-}C}}$ & $\downarrow$ & 0.29 & + + + + + & $\downarrow$ \\
$Q_{\mathrm{SNR}}$ & $\uparrow$ & $-$0.25 & $-$ $-$ $-$ $-$ $-$ & $\uparrow$ \\
$Q_{\mathrm{IPR}}$ & -- & 0.04 & + + + + $-$ & $\downarrow$ \\
$Q_{\mathrm{BPR}}$ & $\uparrow$ & $-$0.25 & $-$ $-$ $-$ $-$ $-$ & $\uparrow$ \\
$Q_{\mathrm{KURT}}$ & -- & $-$0.12 & $-$ $-$ + $-$ $-$ & $\uparrow$ \\
$Q_{\mathrm{SKEW}}$ & -- & 0.18 & + + + + $-$ & $\downarrow$ \\
$Q_{\mathrm{PerI}}$ & -- & $-$0.10 & $-$ $-$ $-$ $-$ + & $\uparrow$ \\
$Q_{\mathrm{TMCC}}$ & $\uparrow$ & $-$0.27 & $-$ $-$ $-$ $-$ $-$ & $\uparrow$ \\
$Q_{\mathrm{HFD}}$ & $\downarrow$ & 0.23 & + + + + + & $\downarrow$ \\
$Q_{\mathrm{KFD}}$ & -- & $-$0.12 & $-$ $-$ $-$ $-$ + & $\uparrow$ \\
\multicolumn{5}{l}{\textit{RR}} \\
$Q_{\mathrm{ZCR}}$ & -- & 0.00 & + $-$ + $-$ $-$ & $\downarrow$ \\
$Q_{\mathrm{HJ\text{-}M}}$ & -- & $-$0.01 & + $-$ + $-$ $-$ & $\downarrow$ \\
$Q_{\mathrm{HJ\text{-}C}}$ & -- & 0.05 & $-$ + $-$ + + & $\uparrow$ \\
$Q_{\mathrm{SNR}}$ & $\uparrow$ & $-$0.12 & $-$ $-$ $-$ $-$ $-$ & $\uparrow$ \\
$Q_{\mathrm{IPR}}$ & -- & 0.01 & + + $-$ + $-$ & $\downarrow$ \\
$Q_{\mathrm{BPR}}$ & $\uparrow$ & $-$0.12 & $-$ $-$ $-$ $-$ $-$ & $\uparrow$ \\
$Q_{\mathrm{KURT}}$ & $\uparrow$ & $-$0.07 & $-$ $-$ $-$ $-$ $-$ & -- \\
$Q_{\mathrm{SKEW}}$ & $\downarrow$ & 0.05 & + + + + + & $\uparrow$ \\
$Q_{\mathrm{PerI}}$ & -- & $-$0.05 & $-$ $-$ + $-$ $-$ & $\downarrow$ \\
$Q_{\mathrm{TMCC}}$ & -- & 0.03 & + + + $-$ $-$ & $\downarrow$ \\
$Q_{\mathrm{HFD}}$ & $\downarrow$ & 0.03 & + + + + + & $\downarrow$ \\
$Q_{\mathrm{KFD}}$ & -- & $-$0.04 & $-$ $-$ + $-$ $-$ & $\downarrow$ \\
\bottomrule
\end{tabular}
\par\smallskip\begin{minipage}{\linewidth}\footnotesize\textit{Note.} $\rho_{\times}$ is the raw Spearman correlation between indicator and absolute error across methods within each window, averaged per subject (negative = higher is better); it is the quantity the direction rule uses. Within-method and partial correlations for the Trainset are given in Section~\ref{sec:s6} (median over methods, oriented). Subject signs: sign of the cross-method $\rho$ in each of the five subjects; an indicator is kept only if all five agree. Label-free: direction inferred with the error measured against the median of the other methods (a pseudo-reference) instead of the contact reference; it comes from its own correlation, not from the $\rho_{\times}$ shown, so the two can differ when $|\rho_{\times}|$ is small. $Q_{\mathrm{BPR}}$ ranks windows exactly as $Q_{\mathrm{SNR}}$ and is excluded from fusion.\end{minipage}
\end{table}

\begin{table}[!htp]
\centering
\caption{Choices fixed on the Trainset.}
\label{tab:s5choices}
\footnotesize
\begin{tabular}{@{}llll@{}}
\toprule
Group & Parameter & HR & RR \\
\midrule
\multirow{3}{*}{Global}
 & Indicator pool & $Q_{\mathrm{HJ\text{-}M}}, Q_{\mathrm{HJ\text{-}C}}, Q_{\mathrm{SNR}}, Q_{\mathrm{TMCC}}, Q_{\mathrm{HFD}}$ & $Q_{\mathrm{SNR}}, Q_{\mathrm{KURT}}, Q_{\mathrm{SKEW}}, Q_{\mathrm{HFD}}$ \\
 & DN1 method & POS + PF & Mot-S + PF \\
 & Constant (median) & 78.95 & 12.52 \\
\midrule
\multirow{4}{*}{ALL}
 & SMM $Q_{\mathrm{H}}$ / $Q_{\mathrm{S}}$ & $Q_{\mathrm{HJ\text{-}C}}, Q_{\mathrm{TMCC}}$ \; (both) & $Q_{\mathrm{SNR}}, Q_{\mathrm{KURT}}, Q_{\mathrm{HFD}}$ \; / \; $Q_{\mathrm{SNR}}, Q_{\mathrm{SKEW}}$ \\
 & SMM $S_{\mathrm{H}}$ / $S_{\mathrm{S}}$ & $S_{\mathrm{NCC}}, S_{\mathrm{PFD}}$ \; / \; $S_{\mathrm{PFD}}, S_{\mathrm{BPC}}$ & $S_{\mathrm{PFD}}, S_{\mathrm{BPC}}$ \; / \; $S_{\mathrm{NCC}}, S_{\mathrm{PFD}}$ \\
 & $\tau^{*}$ ($Q$ / $S$) & 0.2 \; / \; 0.5 & 0.1 \; / \; 0.1 \\
\midrule
\multirow{3}{*}{NLM}
 & SMM $Q_{\mathrm{H}}, Q_{\mathrm{S}}$ & $Q_{\mathrm{HJ\text{-}C}}, Q_{\mathrm{TMCC}}$ \; (both) & $Q_{\mathrm{SNR}}, Q_{\mathrm{SKEW}}$ \; (both) \\
 & SMM $S_{\mathrm{H}}$ / $S_{\mathrm{S}}$ & $S_{\mathrm{PFD}}, S_{\mathrm{BPC}}$ \; (both) & $S_{\rho}, S_{\mathrm{NCC}}, S_{\cos}$ \; / \; $S_{\rho}, S_{\cos}$ \\
 & $\tau^{*}$ ($Q$ / $S$) & 0.5 \; / \; 0.5 & 0.2 \; / \; 0.75 \\
\midrule
\multirow{3}{*}{DLM}
 & SMM $Q_{\mathrm{H}}, Q_{\mathrm{S}}$ & $Q_{\mathrm{HJ\text{-}C}}, Q_{\mathrm{TMCC}}$ \; (both) & $Q_{\mathrm{SNR}}, Q_{\mathrm{KURT}}$ \; (both) \\
 & SMM $S_{\mathrm{H}}$ / $S_{\mathrm{S}}$ & $S_{\mathrm{MAE}}, S_{\mathrm{BPC}}, S_{\mathrm{BEC}}$ \; / \; $S_{\mathrm{PFD}}, S_{\mathrm{BEC}}$ & $S_{\mathrm{PFD}}, S_{\mathrm{BPC}}$ \; (both) \\
 & $\tau^{*}$ ($Q$ / $S$) & $\infty$ \; / \; $\infty$ & 0.2 \; / \; 0.2 \\
\midrule
\multirow{2}{*}{DN1}
 & SMM $Q_{\mathrm{H}}$ / $Q_{\mathrm{S}}$ & $Q_{\mathrm{HJ\text{-}C}}, Q_{\mathrm{SNR}}, Q_{\mathrm{HFD}}$ \; / \; $Q_{\mathrm{HJ\text{-}C}}, Q_{\mathrm{HFD}}$ & $Q_{\mathrm{SNR}}, Q_{\mathrm{KURT}}$ \; / \; $Q_{\mathrm{KURT}}, Q_{\mathrm{HFD}}$ \\
 & $\tau^{*}$ ($Q$) & 0.5 & 1.0 \\
\bottomrule
\end{tabular}
\par\smallskip\begin{minipage}{\linewidth}\footnotesize\textit{Note.} SMM subsets are fixed on the Trainset per score type ($Q$ quality, $S$ similarity; subscripts $\mathrm{H}$ hard, $\mathrm{S}$ soft) and method group; ``(both)'' indicates identical subsets for hard and soft scores. $\tau^{*}$ is the optimal fusion temperature.\end{minipage}
\end{table}

\begin{table}[!htp]
\centering
\caption{Learned ridge coefficients.}
\label{tab:s5ridge}
\footnotesize
\begin{tabular}{@{}llrrcc@{}}
\toprule
Task & Feature & L-Q & L-Q+rate & Direction & Agrees \\
\midrule
\multirow{5}{*}{RR}
 & $Q_{\mathrm{SNR}}$ & $-$0.120 & $-$0.127 & $\uparrow$ & yes \\
 & $Q_{\mathrm{KURT}}$ & $-$0.038 & $-$0.035 & $\uparrow$ & yes \\
 & $Q_{\mathrm{SKEW}}$ & 0.025 & 0.029 & $\downarrow$ & yes \\
 & $Q_{\mathrm{HFD}}$ & $-$0.002 & 0.000 & $\downarrow$ & no \\
 & predicted rate & — & 0.043 & — & — \\
\midrule
\multirow{6}{*}{HR}
 & $Q_{\mathrm{HJ\text{-}M}}$ & $-$0.076 & $-$0.041 & $\downarrow$ & no \\
 & $Q_{\mathrm{HJ\text{-}C}}$ & $-$0.024 & $-$0.049 & $\downarrow$ & no \\
 & $Q_{\mathrm{SNR}}$ & 0.018 & 0.018 & $\uparrow$ & no \\
 & $Q_{\mathrm{TMCC}}$ & $-$0.186 & $-$0.178 & $\uparrow$ & yes \\
 & $Q_{\mathrm{HFD}}$ & 0.297 & 0.295 & $\downarrow$ & yes \\
 & predicted rate & — & $-$0.050 & — & — \\
\bottomrule
\end{tabular}
\par\smallskip\begin{minipage}{\linewidth}\footnotesize\textit{Note.} Standardised features; positive = higher predicted error. Both models selected $\alpha = 1000$ by leave-one-subject-out validation. Agrees: coefficient sign matches the Trainset direction.\end{minipage}
\end{table}

\section{Fusion temperature and weight sharpness}\label{sec:s4}
\textit{Supports main-text Sections~3.4 and~4.3.} How the temperature was tuned on the Trainset and how sharply it concentrates the weights.

\begin{figure}[!htp]\centering\includegraphics[width=\linewidth]{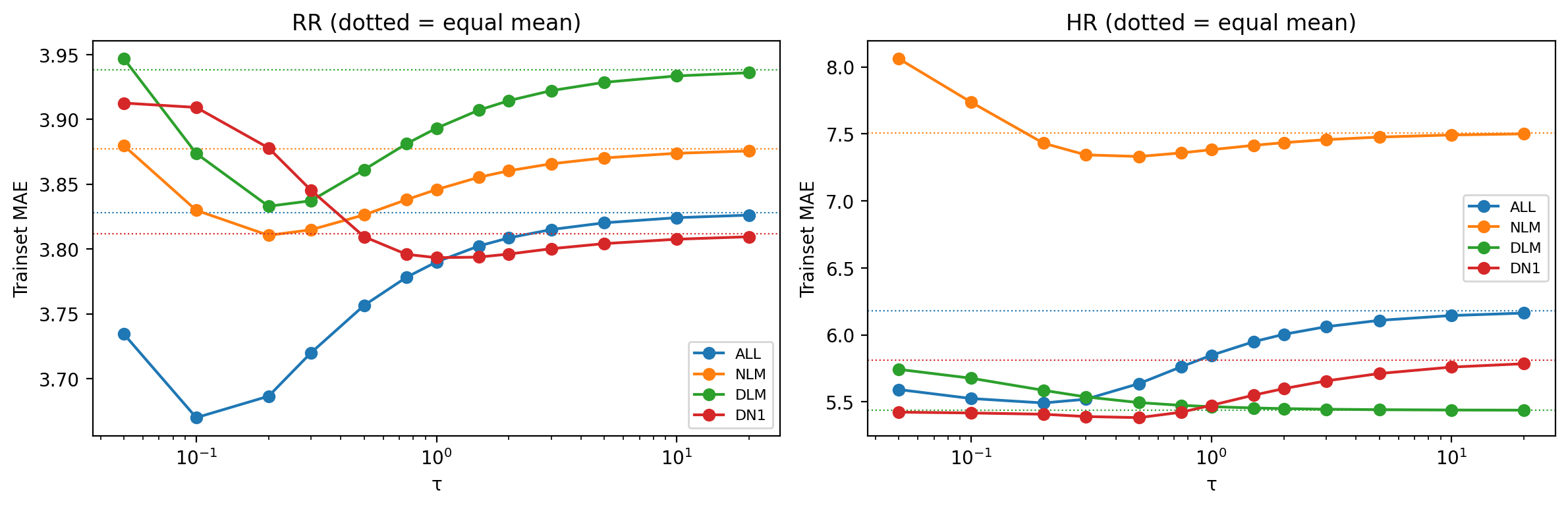}\caption{Trainset MAE of quality-weighted soft fusion against the temperature $\tau$, per group. Dotted lines: equal mean ($\tau=\infty$); the minimum defines $\tau^{*}$.}\label{fig:s7tau}\end{figure}

The curves show where the tuned temperatures come from and how much they promised on the Trainset. For HR (ALL), $\tau^{*}=0.2$ lowered Trainset MAE from 6.18 to 5.49\,bpm (11.2\%), and the test sets reproduced this (10.9\% [5.2, 18.4] over the equal mean, main-text Table~4). For HR DLM the curve falls monotonically toward the equal mean, so $\tau^{*}=\infty$. For RR (ALL), $\tau^{*}=0.1$ lowered Trainset MAE from 3.83 to 3.67\,brpm (4.1\%), but on the test sets the same temperature increased error ($-2.6$\% [$-4.8$, $-0.8$]). The RR minimum is narrow (MAE rises again at $\tau=0.05$), and the other RR groups gain only 0.02--0.11\,brpm over the equal mean even at their own optima. The RR temperature therefore fits the five Trainset subjects, whereas the HR temperature captures quality information that transfers.

\begin{table}[!htp]
\centering
\caption{Weight sharpness of quality-weighted soft fusion.}
\label{tab:s7sharp}
\footnotesize
\begin{tabular}{@{}llclcc@{}}
\toprule
Task & Group & Methods & Setting & Largest weight & Effective methods \\
\midrule
\multirow{8}{*}{HR}
 & \multirow{2}{*}{ALL} & \multirow{2}{*}{7} & $\tau = 1$ & 0.21 & 6.25--6.33 \\
 &  &  & $\tau^{*} = 0.20$ & 0.42--0.43 & 2.99--3.09 \\
\cmidrule(lr){2-6}
 & \multirow{2}{*}{NLM} & \multirow{2}{*}{4} & $\tau = 1$ & 0.30--0.31 & 3.83--3.88 \\
 &  &  & $\tau^{*} = 0.50$ & 0.36--0.38 & 3.47--3.61 \\
\cmidrule(lr){2-6}
 & \multirow{2}{*}{DLM} & \multirow{2}{*}{3} & $\tau = 1$ & 0.40--0.42 & 2.82--2.88 \\
 &  &  & $\tau^{*} = \infty$ & 0.33 & 3.00 \\
\cmidrule(lr){2-6}
 & \multirow{2}{*}{DN1} & \multirow{2}{*}{2} & $\tau = 1$ & 0.71--0.72 & 1.68--1.69 \\
 &  &  & $\tau^{*} = 0.50$ & 0.85--0.86 & 1.32--1.33 \\
\midrule
\multirow{8}{*}{RR}
 & \multirow{2}{*}{ALL} & \multirow{2}{*}{10} & $\tau = 1$ & 0.13 & 9.67--9.70 \\
 &  &  & $\tau^{*} = 0.10$ & 0.48--0.49 & 3.38--3.50 \\
\cmidrule(lr){2-6}
 & \multirow{2}{*}{NLM} & \multirow{2}{*}{5} & $\tau = 1$ & 0.25--0.26 & 4.79--4.81 \\
 &  &  & $\tau^{*} = 0.20$ & 0.47--0.49 & 2.99--3.10 \\
\cmidrule(lr){2-6}
 & \multirow{2}{*}{DLM} & \multirow{2}{*}{5} & $\tau = 1$ & 0.26 & 4.77--4.79 \\
 &  &  & $\tau^{*} = 0.20$ & 0.50--0.51 & 2.84--2.93 \\
\cmidrule(lr){2-6}
 & \multirow{2}{*}{DN1} & \multirow{2}{*}{2} & $\tau = 1$ & 0.61--0.62 & 1.86--1.87 \\
 &  &  & $\tau^{*} = 1.00$ & 0.61--0.62 & 1.86--1.87 \\
\bottomrule
\end{tabular}
\par\smallskip\begin{minipage}{\linewidth}\footnotesize\textit{Note.} Mean largest weight and mean effective number of methods ($1/\sum_i w_i^2$); range over the four test sets.\end{minipage}
\end{table}

\FloatBarrier
\bigskip\noindent{\large\bfseries B. Additional results}\par\medskip

\section{Single-method results}\label{sec:s5}
\textit{Supports main-text Section~4.1 and Fig.~2.} MAE and Pearson $r$ for every method and rate estimator. MUSIC had the lowest mean MAE on the Trainset for both tasks and is used throughout.

\begin{table}[!htp]
\centering
\caption{HR: MAE (bpm) of every method and rate estimator.}
\label{tab:s3hr}
\footnotesize
\begin{adjustbox}{max width=\linewidth}
\begin{tabular}{llccccc}
\toprule
Method & Estimator & Trainset & OMS-Test & COHFACE & MAHNOB & StressID \\
\midrule
CHROM & FFT & 9.28 (0.40) & 11.45 (0.37) & 10.58 (0.28) & 13.73 (0.16) & 23.03 (0.00) \\
 & Welch & 10.41 (0.33) & 12.96 (0.31) & 12.07 (0.25) & 16.06 (0.08) & 24.48 (0.01) \\
 & MUSIC & 8.39 (0.46) & 10.06 (0.43) & 8.96 (0.31) & 12.85 (0.15) & 20.05 (0.01) \\
 & Peak & 11.56 (0.20) & 10.70 (0.22) & 10.36 (0.20) & 13.05 ($-$0.03) & 16.40 (0.09) \\
\addlinespace[1pt]
GREEN & FFT & 13.41 (0.17) & 14.31 (0.16) & 9.07 (0.36) & 14.55 (0.05) & 19.44 (0.08) \\
 & Welch & 14.34 (0.15) & 15.68 (0.12) & 10.97 (0.30) & 16.48 (0.06) & 20.21 (0.07) \\
 & MUSIC & 12.36 (0.18) & 12.77 (0.18) & 7.80 (0.44) & 13.54 (0.04) & 18.81 (0.08) \\
 & Peak & 13.05 (0.11) & 12.52 (0.11) & 9.76 (0.23) & 13.78 ($-$0.01) & 17.61 (0.10) \\
\addlinespace[1pt]
OMIT & FFT & 8.75 (0.39) & 10.17 (0.38) & 9.86 (0.31) & 13.01 (0.20) & 23.33 (0.02) \\
 & Welch & 9.56 (0.36) & 11.62 (0.33) & 11.36 (0.26) & 15.22 (0.14) & 24.62 (0.03) \\
 & MUSIC & 8.38 (0.42) & 9.25 (0.41) & 8.31 (0.36) & 11.62 (0.24) & 19.81 (0.05) \\
 & Peak & 10.84 (0.17) & 9.96 (0.25) & 9.55 (0.26) & 11.94 (0.10) & 17.30 (0.02) \\
\addlinespace[1pt]
POS & FFT & 8.18 (0.46) & 9.46 (0.45) & 9.80 (0.32) & 12.71 (0.19) & 24.28 (0.04) \\
 & Welch & 8.97 (0.41) & 11.13 (0.39) & 12.03 (0.25) & 15.57 (0.13) & 25.51 (0.03) \\
 & MUSIC & 7.62 (0.50) & 8.41 (0.50) & 8.73 (0.33) & 11.90 (0.21) & 20.34 (0.04) \\
 & Peak & 9.70 (0.31) & 9.72 (0.31) & 10.05 (0.24) & 11.31 (0.19) & 17.07 ($-$0.04) \\
\midrule
PF & FFT & 5.84 (0.66) & 6.09 (0.69) & 8.76 (0.35) & 11.40 (0.06) & 13.39 (0.06) \\
 & Welch & 6.20 (0.66) & 7.02 (0.62) & 9.77 (0.28) & 12.50 (0.04) & 14.19 (0.05) \\
 & MUSIC & 5.42 (0.69) & 6.01 (0.70) & 9.11 (0.41) & 10.91 (0.07) & 13.07 (0.06) \\
 & Peak & 7.30 (0.53) & 7.61 (0.50) & 8.88 (0.31) & 10.71 (0.14) & 14.66 (0.00) \\
\addlinespace[1pt]
CP$^{+}$ & FFT & 6.37 (0.57) & 5.95 (0.64) & 7.61 (0.40) & 7.44 (0.49) & 15.28 ($-$0.04) \\
 & Welch & 6.81 (0.56) & 6.60 (0.60) & 8.55 (0.33) & 9.07 (0.36) & 15.54 ($-$0.01) \\
 & MUSIC & 5.92 (0.61) & 5.59 (0.66) & 7.29 (0.51) & 7.29 (0.49) & 14.66 ($-$0.01) \\
 & Peak & 7.77 (0.51) & 6.37 (0.55) & 8.13 (0.38) & 8.16 (0.49) & 16.57 ($-$0.05) \\
\addlinespace[1pt]
CP & FFT & 6.47 (0.59) & 5.90 (0.68) & 7.38 (0.43) & 6.11 (0.60) & 14.65 (0.11) \\
 & Welch & 6.86 (0.57) & 6.68 (0.62) & 8.57 (0.35) & 7.31 (0.53) & 15.18 (0.11) \\
 & MUSIC & 5.92 (0.62) & 5.46 (0.70) & 7.48 (0.50) & 5.99 (0.62) & 14.03 (0.13) \\
 & Peak & 7.50 (0.49) & 6.67 (0.56) & 8.27 (0.33) & 7.81 (0.55) & 16.51 (0.03) \\
\bottomrule
\end{tabular}\end{adjustbox}
\par\smallskip\begin{minipage}{\linewidth}\footnotesize\textit{Note.} Pearson $r$ in parentheses. MUSIC (lowest Trainset mean MAE) is used in the main analysis.\end{minipage}
\end{table}

\begin{table}[!htp]
\centering
\caption{RR: MAE (brpm) of every method and rate estimator.}
\label{tab:s3rr}
\footnotesize
\begin{adjustbox}{max width=\linewidth}
\begin{tabular}{llccccc}
\toprule
Method & Estimator & Trainset & OMS-Test & COHFACE & MAHNOB & StressID \\
\midrule
CHROM & FFT & 5.01 (0.03) & 5.36 (0.08) & 4.87 ($-$0.05) & 7.09 (0.02) & 6.28 (0.05) \\
 & Welch & 5.05 (0.05) & 5.52 (0.08) & 5.00 ($-$0.06) & 7.12 (0.01) & 6.26 (0.06) \\
 & MUSIC & 4.67 (0.07) & 5.12 (0.07) & 4.16 ($-$0.06) & 6.72 ($-$0.01) & 5.84 (0.07) \\
 & Peak & 5.44 (0.11) & 6.71 ($-$0.01) & 5.91 ($-$0.04) & 6.25 ($-$0.05) & 6.22 (0.03) \\
\addlinespace[1pt]
GREEN & FFT & 4.44 (0.16) & 5.12 (0.10) & 4.75 ($-$0.03) & 7.11 (0.00) & 6.45 (0.00) \\
 & Welch & 4.63 (0.14) & 5.32 (0.10) & 4.86 ($-$0.04) & 7.08 (0.00) & 6.51 (0.01) \\
 & MUSIC & 4.20 (0.18) & 4.88 (0.11) & 4.05 ($-$0.05) & 6.58 (0.00) & 5.97 ($-$0.01) \\
 & Peak & 5.83 (0.07) & 6.22 (0.05) & 5.81 ($-$0.02) & 6.13 ($-$0.09) & 6.32 (0.03) \\
\addlinespace[1pt]
OMIT & FFT & 4.86 (0.11) & 5.27 (0.10) & 4.83 ($-$0.05) & 7.26 (0.00) & 6.18 (0.08) \\
 & Welch & 4.96 (0.07) & 5.35 (0.13) & 4.95 ($-$0.07) & 7.28 ($-$0.01) & 6.25 (0.07) \\
 & MUSIC & 4.61 (0.11) & 5.02 (0.11) & 4.16 ($-$0.07) & 6.78 (0.01) & 5.84 (0.08) \\
 & Peak & 5.43 (0.08) & 6.26 (0.07) & 5.85 ($-$0.01) & 5.82 (0.01) & 6.21 (0.04) \\
\addlinespace[1pt]
Mot-F & FFT & 4.83 (0.15) & 5.11 (0.15) & 5.15 ($-$0.02) & 6.53 ($-$0.01) & 5.81 (0.13) \\
 & Welch & 5.16 (0.12) & 5.42 (0.15) & 5.35 ($-$0.02) & 6.54 (0.02) & 5.99 (0.09) \\
 & MUSIC & 4.37 (0.17) & 4.72 (0.19) & 4.43 ($-$0.05) & 5.96 (0.02) & 5.32 (0.14) \\
 & Peak & 6.68 (0.05) & 7.15 (0.03) & 6.48 (0.04) & 5.81 (0.00) & 6.39 (0.05) \\
\addlinespace[1pt]
Mot-S & FFT & 4.24 (0.25) & 4.92 (0.18) & 4.90 ($-$0.06) & 6.46 (0.07) & 6.29 (0.04) \\
 & Welch & 4.69 (0.18) & 5.21 (0.15) & 5.08 ($-$0.05) & 6.56 (0.07) & 6.46 (0.01) \\
 & MUSIC & 4.01 (0.25) & 4.64 (0.16) & 4.27 ($-$0.07) & 6.01 (0.06) & 5.69 (0.07) \\
 & Peak & 6.39 (0.07) & 6.58 (0.05) & 6.28 (0.00) & 5.82 (0.01) & 6.47 (0.00) \\
\midrule
EP & FFT & 4.52 (0.12) & 5.41 ($-$0.05) & 4.52 (0.03) & 6.62 (0.03) & 6.28 (0.02) \\
 & Welch & 4.72 (0.10) & 5.60 ($-$0.05) & 4.67 (0.02) & 6.75 ($-$0.01) & 6.28 (0.01) \\
 & MUSIC & 4.26 (0.12) & 5.05 ($-$0.03) & 3.77 (0.03) & 6.33 (0.00) & 5.75 (0.03) \\
 & Peak & 6.20 (0.04) & 6.62 (0.01) & 5.93 ($-$0.01) & 5.74 ($-$0.02) & 6.54 ($-$0.01) \\
\addlinespace[1pt]
IN & FFT & 4.79 (0.08) & 5.21 (0.05) & 4.82 (0.03) & 5.47 (0.05) & 5.51 (0.08) \\
 & Welch & 5.06 (0.05) & 5.32 (0.07) & 4.94 (0.02) & 5.46 (0.05) & 5.64 (0.07) \\
 & MUSIC & 4.44 (0.08) & 4.89 (0.07) & 4.12 (0.02) & 5.24 (0.02) & 5.27 (0.06) \\
 & Peak & 6.09 (0.01) & 6.24 (0.09) & 5.61 (0.02) & 5.06 (0.07) & 6.22 ($-$0.02) \\
\addlinespace[1pt]
PF & FFT & 4.43 (0.11) & 4.85 (0.10) & 4.52 ($-$0.05) & 6.42 ($-$0.04) & 5.95 (0.03) \\
 & Welch & 4.61 (0.11) & 5.01 (0.11) & 4.62 ($-$0.05) & 6.33 (0.00) & 6.13 (0.00) \\
 & MUSIC & 4.13 (0.13) & 4.55 (0.13) & 3.78 ($-$0.05) & 6.00 ($-$0.03) & 5.54 (0.02) \\
 & Peak & 5.80 (0.03) & 6.12 (0.02) & 5.70 (0.00) & 5.39 (0.04) & 6.07 (0.09) \\
\addlinespace[1pt]
CP$^{+}$ & FFT & 4.75 (0.12) & 5.69 (0.09) & 6.08 ($-$0.06) & 6.37 ($-$0.09) & 6.32 (0.08) \\
 & Welch & 5.16 (0.09) & 6.03 (0.09) & 6.27 ($-$0.05) & 6.62 ($-$0.09) & 6.51 (0.08) \\
 & MUSIC & 4.52 (0.13) & 5.31 (0.12) & 5.41 ($-$0.07) & 5.78 ($-$0.10) & 5.59 (0.09) \\
 & Peak & 6.63 (0.03) & 7.10 (0.08) & 7.14 (0.02) & 6.07 ($-$0.04) & 6.75 (0.01) \\
\addlinespace[1pt]
CP & FFT & 5.03 (0.01) & 5.06 (0.13) & 4.70 ($-$0.05) & 6.48 ($-$0.04) & 5.95 (0.04) \\
 & Welch & 5.20 ($-$0.01) & 5.23 (0.10) & 4.87 ($-$0.05) & 6.52 ($-$0.04) & 6.01 (0.03) \\
 & MUSIC & 4.70 (0.00) & 4.83 (0.12) & 4.03 ($-$0.06) & 6.05 ($-$0.04) & 5.48 (0.05) \\
 & Peak & 6.27 ($-$0.03) & 6.43 (0.07) & 6.04 ($-$0.01) & 6.00 ($-$0.06) & 6.40 (0.01) \\
\bottomrule
\end{tabular}\end{adjustbox}
\par\smallskip\begin{minipage}{\linewidth}\footnotesize\textit{Note.} Pearson $r$ in parentheses. MUSIC (lowest Trainset mean MAE) is used in the main analysis.\end{minipage}
\end{table}

\begin{table}[!htp]
\centering
\caption{Mean single-method MAE per rate estimator.}
\label{tab:s3means}
\footnotesize
\begin{tabular}{@{}llrrrrr@{}}
\toprule
Task & Estimator & Trainset & OMS-Test & COHFACE & MAHNOB & StressID \\
\midrule
\multirow{4}{*}{HR} & FFT & 8.33 & 9.05 & 9.01 & 11.28 & 19.06 \\
 & Welch & 9.02 & 10.24 & 10.47 & 13.17 & 19.96 \\
 & MUSIC & 7.72 & 8.22 & 8.24 & 10.58 & 17.25 \\
 & Peak & 9.67 & 9.08 & 9.28 & 10.96 & 16.59 \\
\midrule
\multirow{4}{*}{RR} & FFT & 4.69 & 5.20 & 4.91 & 6.58 & 6.10 \\
 & Welch & 4.92 & 5.40 & 5.06 & 6.63 & 6.20 \\
 & MUSIC & 4.39 & 4.90 & 4.22 & 6.14 & 5.63 \\
 & Peak & 6.08 & 6.55 & 6.07 & 5.81 & 6.36 \\
\bottomrule
\end{tabular}
\end{table}

\section{Quality indicators and estimation error}\label{sec:s6}
\textit{Supports main-text Section~4.2 and Figs.~5--6.} For HR the indicators' link to error survives controlling for the predicted rate; for RR it vanishes on MAHNOB-HCI and is too weak to rank methods on COHFACE.

\begin{table}[!htp]
\centering
\caption{HR: oriented Spearman $\rho$ between indicators and absolute error, per dataset.}
\label{tab:s6hr}
\footnotesize
\begin{adjustbox}{max width=\linewidth}
\begin{tabular}{lccccc}
\toprule
Indicator & Trainset & OMS-Test & COHFACE & MAHNOB & StressID \\
\midrule
$Q_{\mathrm{ZCR}}$ & 0.20 / 0.23 / 0.11 & 0.28 / 0.24 / 0.24 & 0.14 / 0.08 / 0.08 & 0.17 / 0.17 / 0.26 & 0.02 / 0.15 / 0.09 \\
$Q_{\mathrm{HJ\text{-}M}}$ & 0.14 / 0.23 / 0.15 & 0.26 / 0.20 / 0.25 & 0.20 / 0.11 / 0.05 & 0.21 / 0.17 / 0.29 & $-$0.03 / 0.27 / 0.12 \\
$Q_{\mathrm{HJ\text{-}C}}$ & 0.16 / $-$0.07 / 0.29 & $-$0.09 / $-$0.01 / 0.26 & $-$0.17 / $-$0.05 / $-$0.02 & $-$0.04 / $-$0.04 / 0.26 & 0.27 / $-$0.04 / 0.30 \\
$Q_{\mathrm{SNR}}$ & 0.18 / 0.19 / 0.25 & 0.26 / 0.22 / 0.28 & 0.09 / 0.03 / 0.03 & $-$0.01 / 0.07 / 0.28 & 0.19 / 0.21 / 0.22 \\
$Q_{\mathrm{IPR}}$ & $-$0.03 / 0.00 / 0.04 & $-$0.03 / $-$0.03 / 0.05 & 0.00 / 0.01 / 0.01 & $-$0.01 / $-$0.01 / 0.05 & 0.02 / 0.02 / 0.04 \\
$Q_{\mathrm{KURT}}$ & $-$0.17 / $-$0.18 / $-$0.12 & 0.00 / 0.00 / $-$0.11 & $-$0.01 / 0.01 / $-$0.04 & $-$0.01 / 0.01 / 0.00 & $-$0.07 / $-$0.07 / $-$0.02 \\
$Q_{\mathrm{SKEW}}$ & $-$0.03 / $-$0.02 / 0.18 & 0.02 / 0.08 / 0.17 & 0.02 / 0.00 / 0.00 & $-$0.04 / $-$0.02 / 0.17 & 0.10 / 0.06 / 0.15 \\
$Q_{\mathrm{PerI}}$ & $-$0.06 / 0.02 / $-$0.10 & $-$0.09 / $-$0.04 / $-$0.14 & $-$0.05 / $-$0.02 / $-$0.01 & 0.00 / $-$0.02 / $-$0.13 & $-$0.07 / $-$0.07 / $-$0.10 \\
$Q_{\mathrm{TMCC}}$ & 0.40 / 0.35 / 0.27 & 0.27 / 0.30 / 0.28 & 0.08 / 0.10 / 0.03 & 0.02 / 0.06 / 0.26 & 0.10 / 0.07 / 0.24 \\
$Q_{\mathrm{HFD}}$ & 0.26 / 0.30 / 0.23 & 0.29 / 0.28 / 0.30 & 0.22 / 0.15 / 0.06 & 0.10 / 0.15 / 0.30 & 0.15 / 0.22 / 0.21 \\
$Q_{\mathrm{KFD}}$ & $-$0.10 / $-$0.11 / $-$0.12 & $-$0.09 / $-$0.10 / $-$0.10 & 0.01 / $-$0.01 / $-$0.02 & $-$0.01 / 0.00 / $-$0.08 & $-$0.09 / 0.00 / $-$0.06 \\
\bottomrule
\end{tabular}\end{adjustbox}
\par\smallskip\begin{minipage}{\linewidth}\footnotesize\textit{Note.} Entries: median within-method / median within-method partial given the predicted rate / cross-method; $\rho>0$ = useful. Pool indicators are oriented by their Trainset direction; dropped indicators are shown with the lower-is-better orientation, so a negative value means higher values went with lower error.\end{minipage}
\end{table}

\begin{table}[!htp]
\centering
\caption{RR: oriented Spearman $\rho$ between indicators and absolute error, per dataset.}
\label{tab:s6rr}
\footnotesize
\begin{adjustbox}{max width=\linewidth}
\begin{tabular}{lccccc}
\toprule
Indicator & Trainset & OMS-Test & COHFACE & MAHNOB & StressID \\
\midrule
$Q_{\mathrm{ZCR}}$ & 0.04 / 0.07 / 0.00 & 0.00 / 0.04 / 0.00 & 0.25 / 0.11 / 0.12 & $-$0.17 / $-$0.01 / $-$0.23 & $-$0.01 / 0.03 / $-$0.06 \\
$Q_{\mathrm{HJ\text{-}M}}$ & 0.03 / 0.14 / $-$0.01 & 0.04 / 0.09 / 0.00 & 0.33 / 0.19 / 0.14 & $-$0.30 / 0.00 / $-$0.28 & 0.02 / 0.08 / $-$0.07 \\
$Q_{\mathrm{HJ\text{-}C}}$ & 0.11 / 0.12 / 0.05 & 0.10 / 0.08 / 0.05 & $-$0.17 / $-$0.08 / $-$0.07 & 0.27 / 0.03 / 0.28 & 0.06 / 0.00 / 0.11 \\
$Q_{\mathrm{SNR}}$ & 0.20 / 0.20 / 0.12 & 0.16 / 0.13 / 0.09 & 0.14 / 0.12 / 0.11 & 0.09 / 0.03 / 0.10 & 0.05 / 0.05 / 0.08 \\
$Q_{\mathrm{IPR}}$ & 0.03 / 0.04 / 0.01 & 0.04 / 0.05 / 0.02 & 0.02 / 0.05 / 0.07 & 0.01 / 0.01 / $-$0.02 & 0.02 / 0.02 / 0.03 \\
$Q_{\mathrm{KURT}}$ & 0.10 / 0.11 / 0.07 & 0.06 / 0.07 / 0.04 & 0.01 / 0.01 / $-$0.01 & $-$0.07 / $-$0.02 / $-$0.07 & 0.01 / 0.01 / $-$0.01 \\
$Q_{\mathrm{SKEW}}$ & 0.06 / 0.07 / 0.05 & 0.02 / 0.03 / 0.01 & $-$0.02 / $-$0.02 / $-$0.02 & 0.01 / 0.01 / $-$0.01 & 0.00 / 0.00 / $-$0.01 \\
$Q_{\mathrm{PerI}}$ & $-$0.04 / $-$0.03 / $-$0.05 & $-$0.05 / $-$0.04 / $-$0.02 & 0.01 / $-$0.01 / 0.01 & $-$0.07 / 0.01 / $-$0.09 & $-$0.01 / 0.00 / $-$0.04 \\
$Q_{\mathrm{TMCC}}$ & 0.04 / 0.07 / 0.03 & 0.03 / 0.05 / 0.03 & 0.25 / 0.15 / 0.13 & $-$0.10 / 0.03 / $-$0.12 & 0.03 / 0.06 / 0.02 \\
$Q_{\mathrm{HFD}}$ & 0.05 / 0.05 / 0.03 & 0.03 / 0.02 / 0.02 & 0.07 / 0.09 / 0.08 & 0.08 / $-$0.01 / 0.08 & 0.04 / 0.03 / 0.06 \\
$Q_{\mathrm{KFD}}$ & $-$0.03 / $-$0.01 / $-$0.04 & $-$0.03 / $-$0.02 / $-$0.05 & 0.11 / 0.03 / 0.06 & $-$0.13 / 0.01 / $-$0.16 & 0.00 / 0.04 / $-$0.05 \\
\bottomrule
\end{tabular}\end{adjustbox}
\par\smallskip\begin{minipage}{\linewidth}\footnotesize\textit{Note.} Entries: median within-method / median within-method partial given the predicted rate / cross-method; $\rho>0$ = useful. Pool indicators are oriented by their Trainset direction; dropped indicators are shown with the lower-is-better orientation, so a negative value means higher values went with lower error.\end{minipage}
\end{table}

\begin{table}[!htp]
\centering
\caption{Significant within-method correlations after Benjamini--Hochberg correction.}
\label{tab:s6counts}
\footnotesize
\begin{tabular}{lcccc}
\toprule
 & \multicolumn{2}{c}{HR} & \multicolumn{2}{c}{RR} \\
\cmidrule(lr){2-3}\cmidrule(lr){4-5}
Indicator & raw & partial & raw & partial \\
\midrule
$Q_{\mathrm{ZCR}}$ & 2 / 0 & 6 / 0 & 3 / 3 & 9 / 0 \\
$Q_{\mathrm{HJ\text{-}M}}$ & 1 / 0 & 9 / 0 & 2 / 6 & 10 / 0 \\
$Q_{\mathrm{HJ\text{-}C}}$ & 0 / 0 & 3 / 3 & 9 / 1 & 0 / 6 \\
$Q_{\mathrm{SNR}}$ & 3 / 0 & 1 / 0 & 6 / 0 & 9 / 0 \\
$Q_{\mathrm{IPR}}$ & 0 / 0 & 0 / 0 & 2 / 0 & 4 / 0 \\
$Q_{\mathrm{KURT}}$ & 0 / 2 & 0 / 2 & 0 / 3 & 2 / 0 \\
$Q_{\mathrm{SKEW}}$ & 0 / 0 & 1 / 0 & 0 / 0 & 1 / 0 \\
$Q_{\mathrm{PerI}}$ & 0 / 0 & 0 / 0 & 0 / 2 & 0 / 0 \\
$Q_{\mathrm{TMCC}}$ & 1 / 0 & 6 / 0 & 4 / 1 & 9 / 0 \\
$Q_{\mathrm{HFD}}$ & 3 / 0 & 11 / 0 & 8 / 0 & 10 / 0 \\
$Q_{\mathrm{KFD}}$ & 0 / 0 & 0 / 2 & 1 / 4 & 2 / 1 \\
\bottomrule
\end{tabular}
\par\smallskip\begin{minipage}{\linewidth}\footnotesize\textit{Note.} Number of method--dataset cells with a significant correlation (two-sided Wilcoxon signed-rank test on subject-level $\rho$), as useful / harmful, for the raw $\rho$ and for the partial $\rho$ given the predicted rate. Only COHFACE and MAHNOB contribute: with five subjects the smallest two-sided $p$ is 0.0625, and with six (0.031) no cell survives correction. Denominators per indicator: 14 cells for HR (7 methods $\times$ 2 datasets) and 20 for RR (10 $\times$ 2).\end{minipage}
\end{table}

For RR the two datasets diverge: every significant MAHNOB-HCI effect disappears given the predicted rate, whereas on COHFACE most indicators are significant in more methods after partialling (Table~\ref{tab:s6rrds}).

\begin{table}[!htp]
\centering
\caption{RR significance counts by dataset.}
\label{tab:s6rrds}
\footnotesize
\begin{tabular}{lcccccc}
\toprule
 & \multicolumn{3}{c}{COHFACE} & \multicolumn{3}{c}{MAHNOB-HCI} \\
\cmidrule(lr){2-4}\cmidrule(lr){5-7}
Indicator & raw & partial & median partial $\rho$ & raw & partial & median partial $\rho$ \\
\midrule
$Q_{\mathrm{ZCR}}$ & 3 / 0 & 9 / 0 & 0.11 & 0 / 3 & 0 / 0 & $-$0.01 \\
$Q_{\mathrm{HJ\text{-}M}}$ & 2 / 0 & 10 / 0 & 0.19 & 0 / 6 & 0 / 0 & 0.00 \\
$Q_{\mathrm{HJ\text{-}C}}$ & 1 / 1 & 0 / 6 & $-$0.08 & 8 / 0 & 0 / 0 & 0.03 \\
$Q_{\mathrm{SNR}}$ & 5 / 0 & 9 / 0 & 0.12 & 1 / 0 & 0 / 0 & 0.03 \\
$Q_{\mathrm{IPR}}$ & 2 / 0 & 4 / 0 & 0.05 & 0 / 0 & 0 / 0 & 0.01 \\
$Q_{\mathrm{KURT}}$ & 0 / 1 & 2 / 0 & 0.01 & 0 / 2 & 0 / 0 & $-$0.02 \\
$Q_{\mathrm{SKEW}}$ & 0 / 0 & 1 / 0 & $-$0.02 & 0 / 0 & 0 / 0 & 0.01 \\
$Q_{\mathrm{PerI}}$ & 0 / 1 & 0 / 0 & $-$0.01 & 0 / 1 & 0 / 0 & 0.01 \\
$Q_{\mathrm{TMCC}}$ & 4 / 0 & 9 / 0 & 0.15 & 0 / 1 & 0 / 0 & 0.03 \\
$Q_{\mathrm{HFD}}$ & 4 / 0 & 10 / 0 & 0.09 & 4 / 0 & 0 / 0 & $-$0.01 \\
$Q_{\mathrm{KFD}}$ & 1 / 1 & 2 / 1 & 0.03 & 0 / 3 & 0 / 0 & 0.01 \\
\bottomrule
\end{tabular}
\par\smallskip\begin{minipage}{\linewidth}\footnotesize\textit{Note.} Useful / harmful, of 10 methods, for raw and partial within-method $\rho$, with the median partial $\rho$ (oriented, $>0$ = useful).\end{minipage}
\end{table}

\section{Fusion results by method group}\label{sec:s7}
\textit{Supports main-text Section~4.3 and Table~3}, which show only the ALL group.

Groups: ALL (all methods), NLM (classical and motion methods), DLM (deep models), DN1 (best classical plus best deep method on the Trainset; HR POS + PF, RR Mot-S + PF). Similarity rows are undefined for DN1 (two methods always tie). GT-SMM rows choose the indicator subset on the test set and are oracle rows. Rows follow the order of the main-text fusion table; rows absent from the main text (GT-SMM, hard learned scores, similarity SMM) are placed next to their closest counterpart.

\begin{table}[!htp]
\centering
\caption{HR fusion MAE (bpm) for all groups. $^{\dagger}$Uses test labels.}
\label{tab:s4hr}
\scriptsize
\begin{adjustbox}{max width=\linewidth}
\begin{tabular}{lcccccccccccccccc}
\toprule
 & \multicolumn{4}{c}{OMS-Test} & \multicolumn{4}{c}{COHFACE} & \multicolumn{4}{c}{MAHNOB} & \multicolumn{4}{c}{StressID} \\
\cmidrule(lr){2-5}\cmidrule(lr){6-9}\cmidrule(lr){10-13}\cmidrule(lr){14-17}
Row & ALL & NLM & DLM & DN1 & ALL & NLM & DLM & DN1 & ALL & NLM & DLM & DN1 & ALL & NLM & DLM & DN1 \\
\midrule
Baseline$^{\dagger}$ & 5.46 & 8.41 & 5.46 & 6.01 & 7.29 & 7.80 & 7.29 & 8.73 & 5.99 & 11.62 & 5.99 & 10.91 & 13.07 & 18.81 & 13.07 & 13.07 \\
Baseline-train & 6.01 & 8.41 & 6.01 & 6.01 & 9.11 & 8.73 & 9.11 & 9.11 & 10.91 & 11.90 & 10.91 & 10.91 & 13.07 & 20.34 & 13.07 & 13.07 \\
Equal mean & 6.75 & 8.54 & 5.27 & 6.56 & 7.14 & 7.36 & 7.65 & 8.18 & 8.66 & 10.83 & 7.56 & 10.34 & 14.63 & 16.81 & 13.28 & 15.32 \\
Median & 5.97 & 8.58 & 5.20 & 6.56 & 6.87 & 7.57 & 7.30 & 8.18 & 8.18 & 11.30 & 6.80 & 10.34 & 14.12 & 17.66 & 13.60 & 15.32 \\
Constant & 11.73 & 11.73 & 11.73 & 11.73 & 12.26 & 12.26 & 12.26 & 12.26 & 11.29 & 11.29 & 11.29 & 11.29 & 13.13 & 13.13 & 13.13 & 13.13 \\
\addlinespace[3pt]
$Q_{\mathrm{S}}$-AMM ($\tau=1$) & 6.20 & 8.41 & 5.20 & 6.09 & 7.14 & 7.27 & 7.55 & 8.31 & 8.03 & 10.81 & 7.42 & 10.19 & 14.00 & 16.62 & 13.38 & 13.80 \\
$Q_{\mathrm{S}}$-AMM ($\tau^{*}$) & 5.29 & 8.36 & 5.27 & 5.92 & 7.32 & 7.20 & 7.65 & 8.58 & 7.22 & 10.83 & 7.56 & 10.39 & 13.46 & 16.48 & 13.28 & 13.17 \\
Trainset-$Q_{\mathrm{S}}$-SMM & 6.32 & 8.87 & 5.18 & 6.11 & 7.23 & 7.44 & 7.72 & 8.29 & 8.24 & 11.13 & 7.95 & 10.28 & 13.82 & 16.85 & 13.18 & 13.72 \\
$Q_{\mathrm{H}}$-AMM & 5.46 & 9.01 & 5.40 & 5.94 & 7.38 & 7.77 & 7.42 & 9.06 & 7.44 & 12.55 & 7.39 & 10.92 & 14.23 & 18.78 & 14.25 & 13.09 \\
Trainset-$Q_{\mathrm{H}}$-SMM & 5.38 & 10.96 & 5.34 & 6.00 & 8.07 & 8.89 & 8.25 & 9.10 & 9.59 & 13.85 & 9.27 & 10.91 & 13.38 & 19.72 & 13.57 & 13.08 \\
\addlinespace[3pt]
L-Q & 5.32 & 8.25 & 5.26 & 5.85 & 7.18 & 7.15 & 7.35 & 8.46 & 8.02 & 11.14 & 7.65 & 10.37 & 13.34 & 16.77 & 13.33 & 13.35 \\
L-Q+rate & 5.27 & 8.67 & 5.20 & 5.88 & 7.19 & 7.24 & 7.35 & 8.50 & 8.77 & 11.82 & 8.05 & 10.97 & 13.84 & 18.67 & 13.21 & 14.46 \\
L-Q (hard) & 5.61 & 9.06 & 5.58 & 5.97 & 7.31 & 7.57 & 7.33 & 8.72 & 8.57 & 13.01 & 8.41 & 10.92 & 14.07 & 19.00 & 13.98 & 13.28 \\
L-Q+rate (hard) & 5.45 & 9.67 & 5.43 & 5.95 & 7.30 & 7.71 & 7.33 & 8.76 & 10.41 & 13.99 & 9.41 & 11.60 & 14.81 & 21.51 & 13.48 & 14.37 \\
\addlinespace[3pt]
$S_{\mathrm{S}}$-AMM & 6.70 & 8.53 & 5.25 & -- & 7.11 & 7.40 & 7.56 & -- & 8.62 & 10.80 & 7.20 & -- & 14.63 & 16.95 & 13.35 & -- \\
Trainset-$S_{\mathrm{S}}$-SMM & 6.55 & 8.48 & 5.25 & -- & 7.09 & 7.46 & 7.56 & -- & 8.51 & 10.85 & 7.26 & -- & 14.53 & 17.06 & 13.34 & -- \\
$S_{\mathrm{H}}$-AMM & 8.11 & 9.33 & 5.41 & -- & 7.87 & 8.26 & 7.39 & -- & 11.25 & 11.66 & 6.55 & -- & 18.58 & 19.48 & 14.03 & -- \\
Trainset-$S_{\mathrm{H}}$-SMM & 5.99 & 9.22 & 5.50 & -- & 7.40 & 8.21 & 7.41 & -- & 8.98 & 11.73 & 6.56 & -- & 14.13 & 19.31 & 13.83 & -- \\
\addlinespace[3pt]
Oracle$^{\dagger}$ & 2.54 & 4.42 & 3.92 & 4.45 & 2.51 & 3.84 & 5.10 & 5.01 & 2.59 & 6.42 & 4.28 & 5.91 & 6.06 & 9.44 & 9.83 & 9.72 \\
Null oracle$^{\dagger}$ & 3.60 & 5.69 & 6.38 & 8.63 & 3.30 & 5.10 & 6.12 & 7.52 & 3.44 & 5.94 & 6.64 & 8.89 & 6.22 & 9.81 & 9.46 & 11.75 \\
GT-$Q_{\mathrm{S}}$-SMM$^{\dagger}$ & 6.13 & 8.12 & 5.17 & 6.06 & 7.07 & 7.20 & 7.40 & 8.22 & 7.92 & 10.62 & 6.94 & 10.06 & 13.82 & 16.45 & 13.18 & 13.72 \\
GT-$Q_{\mathrm{H}}$-SMM$^{\dagger}$ & 5.25 & 8.71 & 5.23 & 5.92 & 7.23 & 7.59 & 7.25 & 8.66 & 6.59 & 11.83 & 6.45 & 10.29 & 13.38 & 17.62 & 13.57 & 13.04 \\
\bottomrule
\end{tabular}\end{adjustbox}
\end{table}

\begin{table}[!htp]
\centering
\caption{RR fusion MAE (brpm) for all groups. $^{\dagger}$Uses test labels.}
\label{tab:s4rr}
\scriptsize
\begin{adjustbox}{max width=\linewidth}
\begin{tabular}{lcccccccccccccccc}
\toprule
 & \multicolumn{4}{c}{OMS-Test} & \multicolumn{4}{c}{COHFACE} & \multicolumn{4}{c}{MAHNOB} & \multicolumn{4}{c}{StressID} \\
\cmidrule(lr){2-5}\cmidrule(lr){6-9}\cmidrule(lr){10-13}\cmidrule(lr){14-17}
Row & ALL & NLM & DLM & DN1 & ALL & NLM & DLM & DN1 & ALL & NLM & DLM & DN1 & ALL & NLM & DLM & DN1 \\
\midrule
Baseline$^{\dagger}$ & 4.55 & 4.64 & 4.55 & 4.55 & 3.77 & 4.05 & 3.77 & 3.78 & 5.24 & 5.96 & 5.24 & 6.00 & 5.27 & 5.31 & 5.27 & 5.54 \\
Baseline-train & 4.64 & 4.64 & 4.55 & 4.64 & 4.27 & 4.27 & 3.78 & 4.27 & 6.01 & 6.01 & 6.00 & 6.01 & 5.69 & 5.69 & 5.54 & 5.69 \\
Equal mean & 4.33 & 4.41 & 4.42 & 4.36 & 3.03 & 3.23 & 3.21 & 3.58 & 5.65 & 6.02 & 5.43 & 5.71 & 4.99 & 5.27 & 4.93 & 5.31 \\
Median & 4.30 & 4.50 & 4.42 & 4.36 & 3.19 & 3.63 & 3.35 & 3.58 & 5.91 & 6.44 & 5.55 & 5.71 & 5.14 & 5.54 & 5.03 & 5.31 \\
Constant & 4.52 & 4.52 & 4.52 & 4.52 & 2.83 & 2.83 & 2.83 & 2.83 & 5.20 & 5.20 & 5.20 & 5.20 & 4.93 & 4.93 & 4.93 & 4.93 \\
\addlinespace[3pt]
$Q_{\mathrm{S}}$-AMM ($\tau=1$) & 4.31 & 4.40 & 4.41 & 4.35 & 3.03 & 3.23 & 3.21 & 3.56 & 5.63 & 6.01 & 5.42 & 5.69 & 4.98 & 5.26 & 4.91 & 5.32 \\
$Q_{\mathrm{S}}$-AMM ($\tau^{*}$) & 4.39 & 4.42 & 4.48 & 4.35 & 3.27 & 3.42 & 3.35 & 3.56 & 5.58 & 6.03 & 5.42 & 5.69 & 5.08 & 5.32 & 4.98 & 5.32 \\
Trainset-$Q_{\mathrm{S}}$-SMM & 4.31 & 4.38 & 4.40 & 4.37 & 3.04 & 3.24 & 3.20 & 3.59 & 5.61 & 5.99 & 5.45 & 5.71 & 4.98 & 5.27 & 4.91 & 5.33 \\
$Q_{\mathrm{H}}$-AMM & 4.68 & 4.65 & 4.75 & 4.51 & 3.86 & 4.04 & 3.90 & 3.93 & 5.67 & 6.27 & 5.63 & 5.91 & 5.42 & 5.59 & 5.36 & 5.59 \\
Trainset-$Q_{\mathrm{H}}$-SMM & 4.66 & 4.59 & 4.74 & 4.51 & 3.80 & 4.03 & 3.87 & 3.96 & 5.64 & 6.14 & 5.63 & 5.94 & 5.40 & 5.55 & 5.35 & 5.58 \\
\addlinespace[3pt]
L-Q & 4.32 & 4.39 & 4.46 & 4.37 & 3.19 & 3.39 & 3.36 & 3.58 & 5.48 & 6.01 & 5.36 & 5.71 & 5.04 & 5.35 & 4.98 & 5.39 \\
L-Q+rate & 4.31 & 4.47 & 4.37 & 4.33 & 3.01 & 3.31 & 3.12 & 3.48 & 5.85 & 6.28 & 5.60 & 5.79 & 5.12 & 5.44 & 4.99 & 5.37 \\
L-Q (hard) & 4.69 & 4.57 & 4.71 & 4.49 & 3.78 & 4.03 & 3.79 & 3.78 & 5.52 & 6.23 & 5.44 & 5.83 & 5.38 & 5.60 & 5.29 & 5.51 \\
L-Q+rate (hard) & 4.55 & 4.74 & 4.64 & 4.44 & 3.69 & 4.01 & 3.65 & 3.74 & 6.12 & 6.85 & 5.83 & 6.02 & 5.51 & 5.90 & 5.36 & 5.58 \\
\addlinespace[3pt]
$S_{\mathrm{S}}$-AMM & 4.32 & 4.41 & 4.41 & -- & 3.03 & 3.24 & 3.22 & -- & 5.67 & 6.05 & 5.42 & -- & 5.01 & 5.28 & 4.93 & -- \\
Trainset-$S_{\mathrm{S}}$-SMM & 4.31 & 4.40 & 4.40 & -- & 3.04 & 3.23 & 3.21 & -- & 5.69 & 5.99 & 5.46 & -- & 5.01 & 5.26 & 4.95 & -- \\
$S_{\mathrm{H}}$-AMM & 4.43 & 4.63 & 4.56 & -- & 3.67 & 3.99 & 3.74 & -- & 6.12 & 6.45 & 5.62 & -- & 5.44 & 5.69 & 5.29 & -- \\
Trainset-$S_{\mathrm{H}}$-SMM & 4.41 & 4.61 & 4.59 & -- & 3.63 & 4.05 & 3.70 & -- & 6.12 & 6.28 & 5.67 & -- & 5.32 & 5.46 & 5.24 & -- \\
\addlinespace[3pt]
Oracle$^{\dagger}$ & 1.27 & 2.15 & 1.92 & 3.04 & 0.90 & 1.74 & 1.38 & 2.59 & 1.76 & 3.21 & 2.47 & 4.31 & 1.47 & 2.68 & 2.09 & 3.93 \\
Null oracle$^{\dagger}$ & 1.33 & 2.16 & 2.11 & 3.63 & 0.76 & 1.44 & 1.24 & 2.46 & 1.64 & 2.96 & 2.36 & 4.45 & 1.40 & 2.51 & 2.11 & 4.01 \\
GT-$Q_{\mathrm{S}}$-SMM$^{\dagger}$ & 4.31 & 4.38 & 4.40 & 4.34 & 3.02 & 3.22 & 3.19 & 3.54 & 5.59 & 5.96 & 5.38 & 5.64 & 4.97 & 5.24 & 4.91 & 5.29 \\
GT-$Q_{\mathrm{H}}$-SMM$^{\dagger}$ & 4.64 & 4.51 & 4.73 & 4.46 & 3.72 & 3.93 & 3.74 & 3.81 & 5.43 & 5.95 & 5.47 & 5.73 & 5.26 & 5.44 & 5.33 & 5.52 \\
\bottomrule
\end{tabular}\end{adjustbox}
\end{table}

\section{Fusion gains by dataset and dataset weighting}\label{sec:s8}
\textit{Supports main-text Sections~3.4 and~4.3 and Table~4.} Per-dataset bootstrap gains, and a subject-weighted alternative to the pooled estimate.

\begin{table}[!htp]
\centering
\caption{HR: per-dataset subject-bootstrap gains (\%, ALL group).}
\label{tab:s8hr}
\footnotesize
\begin{adjustbox}{max width=\linewidth}
\begin{tabular}{lcccc}
\toprule
Comparison & OMS-Test & COHFACE & MAHNOB & StressID \\
\midrule
Baseline$^{\dagger}$ $\rightarrow$ Equal mean & $-$23.6 [$-$58.5, $-$1.6] & 2.1 [$-$3.0, 6.5] & $-$44.6 [$-$70.0, $-$27.1] & $-$11.9 [$-$49.6, 6.5] \\
Baseline-train $\rightarrow$ Equal mean & $-$12.3 [$-$37.9, 0.2] & 21.6 [6.9, 33.0] & 20.7 [$-$3.6, 34.5] & $-$11.9 [$-$49.6, 6.5] \\
Constant $\rightarrow$ Equal mean & 42.5 [26.2, 58.9] & 41.8 [28.3, 52.6] & 23.3 [5.3, 36.8] & $-$11.4 [$-$38.4, 5.4] \\
Equal mean $\rightarrow$ $Q_{\mathrm{S}}$-AMM ($\tau=1$) & 8.1 [2.5, 13.9] & 0.0 [$-$2.1, 2.2] & 7.2 [3.0, 12.2] & 4.3 [0.7, 10.7] \\
Equal mean $\rightarrow$ $Q_{\mathrm{S}}$-AMM ($\tau^{*}$) & 21.7 [4.6, 40.0] & $-$2.5 [$-$8.1, 3.1] & 16.5 [4.9, 29.7] & 8.0 [$-$2.8, 24.6] \\
Equal mean $\rightarrow$ Trainset-$Q_{\mathrm{S}}$-SMM & 6.3 [2.1, 12.1] & $-$1.2 [$-$4.8, 2.3] & 4.8 [$-$0.8, 11.7] & 5.5 [1.6, 12.7] \\
Equal mean $\rightarrow$ L-Q & 21.2 [4.6, 39.1] & $-$0.6 [$-$4.4, 3.6] & 7.3 [$-$2.5, 18.6] & 8.8 [2.6, 21.0] \\
Equal mean $\rightarrow$ L-Q+rate & 21.8 [6.3, 39.4] & $-$0.7 [$-$4.7, 3.5] & $-$1.4 [$-$15.0, 13.9] & 5.4 [$-$6.8, 16.5] \\
Equal mean $\rightarrow$ $S_{\mathrm{S}}$-AMM & 0.7 [$-$0.7, 2.5] & 0.4 [$-$0.1, 0.9] & 0.4 [$-$0.5, 1.2] & 0.0 [$-$0.6, 0.3] \\
Equal mean $\rightarrow$ $S_{\mathrm{H}}$-AMM & $-$20.2 [$-$26.4, $-$13.3] & $-$10.2 [$-$20.8, 0.4] & $-$30.0 [$-$52.4, $-$11.4] & $-$27.0 [$-$53.2, $-$12.7] \\
Baseline-train $\rightarrow$ $Q_{\mathrm{S}}$-AMM ($\tau^{*}$) & 12.0 [0.3, 26.1] & 19.6 [8.3, 28.7] & 33.8 [22.9, 42.4] & $-$3.0 [$-$25.3, 19.0] \\
Baseline$^{\dagger}$ $\rightarrow$ $Q_{\mathrm{S}}$-AMM ($\tau^{*}$) & 3.2 [0.9, 6.1] & $-$0.4 [$-$7.4, 6.3] & $-$20.7 [$-$37.6, $-$6.2] & $-$3.0 [$-$25.3, 19.0] \\
Baseline-train $\rightarrow$ L-Q & 11.5 [0.2, 23.9] & 21.2 [6.6, 32.7] & 26.5 [14.2, 35.2] & $-$2.0 [$-$20.6, 15.0] \\
Baseline$^{\dagger}$ $\rightarrow$ L-Q & 2.6 [0.8, 4.3] & 1.5 [$-$1.8, 4.8] & $-$34.0 [$-$55.3, $-$17.4] & $-$2.0 [$-$20.6, 15.0] \\
\bottomrule
\end{tabular}\end{adjustbox}
\par\smallskip\begin{minipage}{\linewidth}\footnotesize\textit{Note.} 95\% confidence intervals in brackets; positive favours the second configuration. Subjects: OMS-Test 6, COHFACE 39, MAHNOB 25, StressID 5.\end{minipage}
\end{table}

\begin{table}[!htp]
\centering
\caption{RR: per-dataset subject-bootstrap gains (\%, ALL group).}
\label{tab:s8rr}
\footnotesize
\begin{adjustbox}{max width=\linewidth}
\begin{tabular}{lcccc}
\toprule
Comparison & OMS-Test & COHFACE & MAHNOB & StressID \\
\midrule
Baseline$^{\dagger}$ $\rightarrow$ Equal mean & 4.9 [0.9, 9.4] & 19.8 [11.4, 27.9] & $-$7.7 [$-$16.5, 0.6] & 5.4 [$-$3.0, 12.1] \\
Baseline-train $\rightarrow$ Equal mean & 6.7 [1.1, 13.0] & 29.1 [19.6, 39.0] & 6.0 [0.1, 11.9] & 12.3 [10.9, 13.7] \\
Constant $\rightarrow$ Equal mean & 4.3 [0.6, 7.8] & $-$6.7 [$-$15.8, 2.1] & $-$8.6 [$-$12.3, $-$5.6] & $-$1.2 [$-$6.0, 1.7] \\
Equal mean $\rightarrow$ $Q_{\mathrm{S}}$-AMM ($\tau=1$) & 0.3 [$-$0.1, 0.7] & $-$0.1 [$-$0.5, 0.4] & 0.3 [$-$0.1, 0.7] & 0.2 [$-$0.5, 0.6] \\
Equal mean $\rightarrow$ $Q_{\mathrm{S}}$-AMM ($\tau^{*}$) & $-$1.3 [$-$4.5, 1.7] & $-$8.2 [$-$14.1, $-$4.0] & 1.2 [$-$1.6, 3.9] & $-$1.9 [$-$6.8, 1.8] \\
Equal mean $\rightarrow$ Trainset-$Q_{\mathrm{S}}$-SMM & 0.5 [$-$0.1, 1.0] & $-$0.3 [$-$1.1, 0.3] & 0.6 [0.1, 1.2] & 0.2 [$-$0.8, 0.7] \\
Equal mean $\rightarrow$ L-Q & 0.2 [$-$2.4, 2.4] & $-$5.4 [$-$11.3, $-$1.2] & 2.9 [0.5, 5.4] & $-$1.0 [$-$6.0, 2.7] \\
Equal mean $\rightarrow$ L-Q+rate & 0.4 [$-$1.5, 2.3] & 0.5 [$-$3.3, 4.9] & $-$3.5 [$-$5.4, $-$1.9] & $-$2.6 [$-$3.8, $-$1.4] \\
Equal mean $\rightarrow$ $S_{\mathrm{S}}$-AMM & 0.2 [0.0, 0.6] & $-$0.2 [$-$0.6, 0.3] & $-$0.3 [$-$0.5, $-$0.2] & $-$0.4 [$-$0.6, $-$0.2] \\
Equal mean $\rightarrow$ $S_{\mathrm{H}}$-AMM & $-$2.3 [$-$5.6, 1.4] & $-$21.2 [$-$29.4, $-$14.5] & $-$8.4 [$-$11.7, $-$5.3] & $-$9.0 [$-$11.3, $-$6.8] \\
Baseline-train $\rightarrow$ $Q_{\mathrm{S}}$-AMM ($\tau^{*}$) & 5.4 [0.4, 11.0] & 23.3 [15.4, 31.2] & 7.2 [1.4, 12.7] & 10.6 [5.3, 14.7] \\
Baseline$^{\dagger}$ $\rightarrow$ $Q_{\mathrm{S}}$-AMM ($\tau^{*}$) & 3.6 [$-$2.4, 8.9] & 13.2 [4.4, 20.6] & $-$6.4 [$-$14.8, 1.1] & 3.6 [$-$2.9, 10.9] \\
Baseline-train $\rightarrow$ L-Q & 6.9 [1.8, 12.4] & 25.3 [17.5, 33.2] & 8.8 [3.1, 14.4] & 11.4 [6.1, 15.8] \\
Baseline$^{\dagger}$ $\rightarrow$ L-Q & 5.1 [$-$0.3, 10.0] & 15.4 [5.8, 23.8] & $-$4.6 [$-$12.6, 2.6] & 4.5 [$-$1.3, 11.0] \\
\bottomrule
\end{tabular}\end{adjustbox}
\par\smallskip\begin{minipage}{\linewidth}\footnotesize\textit{Note.} 95\% confidence intervals in brackets; positive favours the second configuration. Subjects: OMS-Test 6, COHFACE 39, MAHNOB 25, StressID 5.\end{minipage}
\end{table}

\begin{table}[!htp]
\centering
\caption{Pooled gains (\%, ALL group) with datasets or subjects weighted equally.}
\label{tab:s8weight}
\footnotesize
\begin{adjustbox}{max width=\linewidth}
\begin{tabular}{lcccc}
\toprule
 & \multicolumn{2}{c}{HR} & \multicolumn{2}{c}{RR} \\
\cmidrule(lr){2-3}\cmidrule(lr){4-5}
Comparison & Equal & Subject & Equal & Subject \\
\midrule
Baseline$^{\dagger}$ $\rightarrow$ Equal mean & $-$19.5 [$-$33.4, $-$10.8] & $-$14.8 [$-$22.3, $-$8.8] & 5.6 [1.9, 9.1] & 6.5 [1.7, 11.0] \\
Baseline-train $\rightarrow$ Equal mean & 4.5 [$-$8.8, 12.0] & 15.2 [2.8, 24.6] & 13.5 [10.5, 16.8] & 16.9 [12.2, 21.5] \\
Constant $\rightarrow$ Equal mean & 24.0 [14.7, 31.4] & 31.6 [21.5, 39.7] & $-$3.1 [$-$6.1, $-$0.4] & $-$6.4 [$-$10.1, $-$2.8] \\
Equal mean $\rightarrow$ $Q_{\mathrm{S}}$-AMM ($\tau=1$) & 4.9 [2.9, 7.6] & 3.9 [2.1, 5.8] & 0.2 [$-$0.1, 0.4] & 0.1 [$-$0.1, 0.4] \\
Equal mean $\rightarrow$ $Q_{\mathrm{S}}$-AMM ($\tau^{*}$) & 10.9 [5.2, 18.4] & 7.9 [2.9, 13.5] & $-$2.6 [$-$4.8, $-$0.8] & $-$2.7 [$-$4.9, $-$0.8] \\
Equal mean $\rightarrow$ Trainset-$Q_{\mathrm{S}}$-SMM & 3.9 [1.7, 6.7] & 2.6 [$-$0.1, 5.4] & 0.2 [$-$0.1, 0.5] & 0.2 [$-$0.2, 0.6] \\
Equal mean $\rightarrow$ L-Q & 9.2 [4.1, 15.7] & 5.3 [1.0, 9.9] & $-$0.8 [$-$2.9, 0.9] & $-$0.8 [$-$3.0, 1.2] \\
Equal mean $\rightarrow$ L-Q+rate & 6.3 [0.3, 13.3] & 2.1 [$-$4.0, 8.4] & $-$1.3 [$-$2.5, $-$0.1] & $-$1.4 [$-$2.8, 0.2] \\
Equal mean $\rightarrow$ $S_{\mathrm{S}}$-AMM & 0.4 [$-$0.1, 0.9] & 0.3 [$-$0.1, 0.7] & $-$0.2 [$-$0.3, 0.0] & $-$0.2 [$-$0.4, 0.0] \\
Equal mean $\rightarrow$ $S_{\mathrm{H}}$-AMM & $-$21.8 [$-$31.0, $-$15.3] & $-$20.4 [$-$29.0, $-$12.1] & $-$10.2 [$-$12.7, $-$8.2] & $-$12.5 [$-$15.3, $-$9.9] \\
Baseline-train $\rightarrow$ $Q_{\mathrm{S}}$-AMM ($\tau^{*}$) & 15.6 [8.1, 23.1] & 22.0 [13.9, 28.7] & 11.6 [8.8, 14.6] & 14.6 [10.7, 18.5] \\
Baseline$^{\dagger}$ $\rightarrow$ $Q_{\mathrm{S}}$-AMM ($\tau^{*}$) & $-$5.2 [$-$12.2, 2.0] & $-$5.7 [$-$12.1, 0.6] & 3.5 [$-$0.1, 6.9] & 4.0 [$-$1.0, 8.2] \\
Baseline-train $\rightarrow$ L-Q & 14.3 [6.8, 20.9] & 19.7 [10.3, 27.4] & 13.1 [10.2, 16.0] & 16.2 [12.4, 20.1] \\
Baseline$^{\dagger}$ $\rightarrow$ L-Q & $-$8.0 [$-$14.7, $-$2.1] & $-$8.7 [$-$14.4, $-$3.2] & 5.1 [1.6, 8.5] & 5.8 [0.8, 10.3] \\
\bottomrule
\end{tabular}\end{adjustbox}
\par\smallskip\begin{minipage}{\linewidth}\footnotesize\textit{Note.} 95\% bootstrap confidence intervals in brackets; positive favours the second configuration. Equal: datasets weighted equally, i.e.\ the mean of the four per-dataset gains, each computed from window-level MAEs (the main analysis; main-text Table~4). Subject: every subject has equal weight (6, 39, 25 and 5 of 75 subjects from OMS-Test, COHFACE, MAHNOB-HCI and StressID); each subject's MAE is computed over its own windows and the gain is $1-\sum_s \mathrm{MAE}_B(s)/\sum_s \mathrm{MAE}_A(s)$ over all 75 subjects. Both columns come from the same 2000 subject-level resamples.\end{minipage}
\end{table}

The subject-weighted gain is not a subject-weighted average of the per-dataset gains: each subject's share of the ratio is proportional to its MAE under $A$, so datasets with large errors (StressID, MAHNOB-HCI) count for more than their subject numbers alone. Subject weighting keeps the sign of every comparison in Table~\ref{tab:s8weight}. It shrinks the HR gains of quality weighting over the equal mean by a fifth to two-thirds. For tuned weighting, COHFACE, where the gain is absent ($-2.5\%$), holds 46\% of the ratio and OMS-Test, where it is largest (21.7\%), falls from a quarter to 7\%; the gain drops from 10.9\% to 7.9\% and remains significant, whereas Trainset-SMM and L-Q+rate lose significance (L-Q+rate also for RR). In return, the equal mean's advantage over the Trainset-chosen HR method becomes significant (15.2\% [2.8, 24.6]): COHFACE and MAHNOB-HCI, where that method transfers poorly (gains of 21.6\% and 20.7\%), hold 86\% of the ratio, while OMS-Test and StressID, where it does better than the equal mean, fall to 14\%.

\section{Alternative respiratory reference}\label{sec:s9}
\textit{Supports main-text Section~5.2.}

The complete pipeline was repeated with peak intervals on the belt as the RR reference (same filters, estimator and indicator candidates; HR unchanged). Every Trainset choice was re-derived under this reference with the procedure of the main analysis: indicator directions and pool, SMM subsets, ridge models, temperatures and the constant predictor. Under this reference, the Trainset kept a larger RR indicator pool ($Q_{\mathrm{ZCR}}$, $Q_{\mathrm{HJ\text{-}M}}$, $Q_{\mathrm{HJ\text{-}C}}$, $Q_{\mathrm{SNR}}$, $Q_{\mathrm{SKEW}}$, $Q_{\mathrm{PerI}}$, $Q_{\mathrm{HFD}}$, $Q_{\mathrm{KFD}}$) and the constant predictor became 14.06\,brpm. Table~\ref{tab:s1peakfusion} gives the fused MAE and Table~\ref{tab:s1peakboot} the pooled bootstrap gains.

\begin{table}[!htp]
\centering
\caption{RR fusion with belt peak intervals as the reference (ALL group).}
\label{tab:s1peakfusion}
\footnotesize
\begin{adjustbox}{max width=\linewidth}
\begin{tabular}{lccccc}
\toprule
Row & OMS-Test & COHFACE & MAHNOB & StressID & Mean \\
\midrule
Baseline$^{\dagger}$ & 4.80 (0.02) & 3.77 (0.03) & 5.26 (0.03) & 5.28 (0.06) & 4.78 \\
Baseline-train & 5.14 (0.04) & 3.78 ($-$0.05) & 6.76 ($-$0.07) & 6.86 (0.02) & 5.64 \\
Equal mean & 5.14 (0.10) & 3.03 ($-$0.09) & 6.27 ($-$0.04) & 6.39 (0.09) & 5.21 \\
Median & 5.39 (0.07) & 3.19 ($-$0.11) & 6.71 ($-$0.06) & 6.77 (0.09) & 5.52 \\
Constant & 4.05 & 3.39 & 4.31 & 5.09 & 4.21 \\
$Q_{\mathrm{S}}$-AMM ($\tau=1$) & 4.91 (0.10) & 3.09 ($-$0.09) & 5.96 ($-$0.03) & 6.11 (0.09) & 5.02 \\
$Q_{\mathrm{S}}$-AMM ($\tau^{*}$) & 4.16 (0.14) & 4.29 ($-$0.03) & 4.36 ($-$0.01) & 4.79 (0.07) & 4.40 \\
Trainset-$Q_{\mathrm{S}}$-SMM & 4.76 (0.11) & 3.23 ($-$0.09) & 5.69 ($-$0.03) & 5.88 (0.09) & 4.89 \\
$Q_{\mathrm{H}}$-AMM & 4.42 (0.12) & 5.40 ($-$0.04) & 4.36 (0.00) & 4.84 (0.06) & 4.76 \\
L-Q & 4.14 (0.14) & 4.12 ($-$0.04) & 4.86 (0.01) & 5.22 (0.09) & 4.58 \\
L-Q+rate & 4.16 (0.16) & 3.78 ($-$0.06) & 4.56 (0.01) & 4.91 (0.10) & 4.35 \\
$S_{\mathrm{S}}$-AMM & 5.16 (0.09) & 3.03 ($-$0.10) & 6.30 ($-$0.04) & 6.43 (0.09) & 5.23 \\
$S_{\mathrm{H}}$-AMM & 5.67 (0.05) & 3.67 ($-$0.09) & 7.00 ($-$0.04) & 7.05 (0.07) & 5.85 \\
Oracle$^{\dagger}$ & 1.85 (0.83) & 0.90 (0.89) & 2.17 (0.66) & 2.29 (0.76) & 1.80 \\
Null oracle$^{\dagger}$ & 1.85 & 0.76 & 2.01 & 2.11 & 1.68 \\
\bottomrule
\end{tabular}\end{adjustbox}
\par\smallskip\begin{minipage}{\linewidth}\footnotesize\textit{Note.} MAE (brpm) with Pearson $r$. COHFACE uses the same rate-series reference in both analyses. $^{\dagger}$Uses test labels.\end{minipage}
\end{table}

\begin{table}[!htp]
\centering
\caption{Pooled bootstrap gains (\%) under the peak RR reference (ALL group, 75 subjects); positive favours the second configuration.}
\label{tab:s1peakboot}
\footnotesize
\begin{tabular}{lc}
\toprule
Comparison & Gain [95\% CI] \\
\midrule
Baseline$^{\dagger}$ $\rightarrow$ Equal mean & $-$6.9 [$-$11.1, $-$3.0] \\
Baseline-train $\rightarrow$ Equal mean & 8.6 [5.5, 11.6] \\
Constant $\rightarrow$ Equal mean & $-$21.8 [$-$27.3, $-$16.7] \\
Equal mean $\rightarrow$ $Q_{\mathrm{S}}$-AMM ($\tau=1$) & 2.8 [1.9, 3.6] \\
Equal mean $\rightarrow$ $Q_{\mathrm{S}}$-AMM ($\tau^{*}$) & 8.1 [$-$1.0, 15.1] \\
Equal mean $\rightarrow$ Trainset-$Q_{\mathrm{S}}$-SMM & 4.5 [2.4, 6.2] \\
Equal mean $\rightarrow$ L-Q & 6.0 [$-$2.4, 12.3] \\
Equal mean $\rightarrow$ L-Q+rate & 11.2 [5.0, 16.0] \\
Equal mean $\rightarrow$ $S_{\mathrm{S}}$-AMM & $-$0.5 [$-$0.6, $-$0.3] \\
Baseline-train $\rightarrow$ $Q_{\mathrm{S}}$-AMM ($\tau^{*}$) & 17.8 [13.0, 21.8] \\
Baseline$^{\dagger}$ $\rightarrow$ $Q_{\mathrm{S}}$-AMM ($\tau^{*}$) & 6.4 [$-$0.7, 12.3] \\
Baseline-train $\rightarrow$ L-Q & 15.7 [11.5, 19.3] \\
Baseline$^{\dagger}$ $\rightarrow$ L-Q & 3.3 [$-$3.5, 8.8] \\
\bottomrule
\end{tabular}
\end{table}

The two belt references agreed only moderately: spectral estimates on the belt compared with peak intervals gave $r$ = 0.35--0.51 and MAE 2.68--4.49\,brpm across datasets and estimators.

\section{RR estimation with 30-s windows}\label{sec:s10}
\textit{Supports main-text Section~5.3.} Longer RR windows improve the estimates and the reference, but not quality-guided fusion, even with $Q_{\mathrm{SNR}}$ restored to the indicator pool.

The RR analysis was repeated with 30-s windows and a 1-s step; HR is unchanged. The RR reference is FFT on the belt over each 30-s window (COHFACE: the mean of the provided rate series over the window). Every Trainset choice was re-derived with the procedure of the main analysis. The Trainset chose Welch as the RR estimator (mean MAE 3.77\,brpm; MUSIC 3.88, FFT 3.99, Peak 4.03), Mot-S + CP$^{+}$ as DN1, and 12.96\,brpm as the constant. $Q_{\mathrm{SNR}}$ (and $Q_{\mathrm{BPR}}$, which ranks windows identically) was dropped because its sign disagreed in one of the five Trainset subjects, leaving the pool $Q_{\mathrm{KURT}}$, $Q_{\mathrm{SKEW}}$, $Q_{\mathrm{PerI}}$. To check that the result did not depend on this, the analysis was repeated with the 10-s pool and directions; $\tau^{*}$ and the ridge regularisation were re-tuned on the 30-s Trainset for every variant.

\begin{table}[!htp]
\centering
\caption{RR in 10-s and 30-s windows. Gains are pooled bootstrap gains (\%, 95\% CI); positive favours the second configuration.}
\label{tab:s9summary}
\footnotesize
\begin{tabular}{lcc}
\toprule
 & 10-s windows & 30-s windows \\
\midrule
Trainset-chosen RR estimator & MUSIC & Welch \\
RR indicator pool & $Q_{\mathrm{SNR}}$, $Q_{\mathrm{KURT}}$, $Q_{\mathrm{SKEW}}$, $Q_{\mathrm{HFD}}$ & $Q_{\mathrm{KURT}}$, $Q_{\mathrm{SKEW}}$, $Q_{\mathrm{PerI}}$ \\
Best single method per test set, mean MAE (brpm) & 4.71 & 4.18 \\
Highest single-method $r$ on a test set & 0.19 & 0.28 \\
Belt references, spectral vs.\ peak: $r$ & 0.35--0.51 & 0.70--0.82 \\
Belt references, spectral vs.\ peak: MAE (brpm) & 2.7--4.5 & 1.5--3.1 \\
$Q_{\mathrm{SNR}}$ within-method $|\rho|$, OMS-Test / MAHNOB & 0.16 / 0.09 & 0.27 / 0.23 \\
Oracle / null oracle, mean MAE (brpm) & 1.35 / 1.28 & 1.24 / 1.26 \\
Constant $\rightarrow$ Equal mean & $-$3.1 [$-$6.1, $-$0.4] & $-$3.3 [$-$8.5, 1.6] \\
Equal mean $\rightarrow$ $Q_{\mathrm{S}}$-AMM ($\tau^{*}$), Trainset pool & $-$2.6 [$-$4.8, $-$0.8] & $-$6.9 [$-$11.5, $-$2.9] \\
\bottomrule
\end{tabular}
\end{table}

\begin{table}[!htp]
\centering
\caption{RR fusion with 30-s windows (ALL group).}
\label{tab:s9fusion}
\scriptsize
\begin{adjustbox}{max width=\linewidth}
\begin{tabular}{lccccccc}
\toprule
Row & OMS-Test & COHFACE & MAHNOB & StressID & Mean & vs.\ equal mean & vs.\ constant \\
\midrule
Equal mean & 3.86 & 2.87 & 5.75 & 4.84 & 4.33 & -- & -- \\
Constant & 3.92 & 2.94 & 4.92 & 4.82 & 4.15 & -- & -- \\
Baseline$^{\dagger}$ & 3.75 & 3.50 & 4.54 & 4.91 & 4.18 & -- & -- \\
Baseline-train & 4.63 & 5.34 & 5.32 & 5.67 & 5.24 & -- & -- \\
\addlinespace[2pt]
\multicolumn{8}{l}{\textit{30-s Trainset pool ($Q_{\mathrm{KURT}}$, $Q_{\mathrm{SKEW}}$, $Q_{\mathrm{PerI}}$)}} \\
AMM ($\tau=1$) & 3.85 & 2.89 & 5.74 & 4.84 & 4.33 & 0.0 [$-$0.5, 0.4] & $-$3.3 [$-$8.4, 1.5] \\
AMM ($\tau^{*}$) & 3.89 & 3.37 & 5.91 & 5.15 & 4.58 & $-$6.9 [$-$11.5, $-$2.9] & $-$10.2 [$-$15.7, $-$5.1] \\
L-Q & 3.84 & 3.00 & 5.77 & 4.88 & 4.37 & $-$1.3 [$-$3.2, 0.5] & $-$4.6 [$-$9.4, 0.0] \\
\addlinespace[2pt]
\multicolumn{8}{l}{\textit{10-s pool ($Q_{\mathrm{SNR}}$, $Q_{\mathrm{KURT}}$, $Q_{\mathrm{SKEW}}$, $Q_{\mathrm{HFD}}$); main check}} \\
AMM ($\tau=1$) & 3.84 & 2.87 & 5.75 & 4.84 & 4.33 & 0.1 [$-$0.2, 0.4] & $-$3.2 [$-$8.5, 1.8] \\
AMM ($\tau^{*}$) & 3.81 & 3.10 & 5.68 & 5.00 & 4.40 & $-$2.2 [$-$5.0, 0.4] & $-$5.4 [$-$10.4, $-$0.8] \\
L-Q & 3.70 & 2.91 & 5.65 & 4.86 & 4.28 & 1.0 [$-$0.8, 2.7] & $-$2.2 [$-$6.7, 2.1] \\
\addlinespace[2pt]
\multicolumn{8}{l}{\textit{Exploratory}} \\
$Q_{\mathrm{SNR}}$ only ($\tau=1$) & 3.80 & 2.88 & 5.73 & 4.84 & 4.31 & 0.4 [$-$0.4, 1.1] & $-$2.9 [$-$8.0, 1.9] \\
$Q_{\mathrm{SNR}}$ only ($\tau^{*}$) & 3.76 & 2.91 & 5.66 & 4.86 & 4.30 & 0.6 [$-$1.0, 2.1] & $-$2.6 [$-$7.3, 1.9] \\
Ridge, all 11 indicators & 3.71 & 2.99 & 5.37 & 4.78 & 4.21 & 1.9 [$-$1.0, 4.4] & $-$1.1 [$-$5.0, 2.6] \\
\bottomrule
\end{tabular}\end{adjustbox}
\par\smallskip\begin{minipage}{\linewidth}\footnotesize\textit{Note.} MAE (brpm) per test set, the unweighted mean over datasets, and pooled bootstrap gains (\%, 95\% CI) over the equal mean and the constant predictor (positive favours the fusion row). $^{\dagger}$Chosen with test labels.\end{minipage}
\end{table}

\begin{table}[!htp]
\centering
\caption{Cross-method $\rho$ between $Q_{\mathrm{SNR}}$ and absolute error with 30-s windows.}
\label{tab:s9snr}
\footnotesize
\begin{tabular}{lccc}
\toprule
Dataset & $\rho_{\times}$ & Subjects $\rho>0$ & $p$ \\
\midrule
Trainset & 0.10 & 4 / 5 & 0.125 \\
OMS-Test & 0.12 & 6 / 6 & 0.031 \\
COHFACE & 0.16 & 35 / 39 & $<$0.001 \\
MAHNOB & 0.09 & 17 / 25 & 0.011 \\
StressID & 0.08 & 4 / 5 & 0.125 \\
\bottomrule
\end{tabular}
\par\smallskip\begin{minipage}{\linewidth}\footnotesize\textit{Note.} Oriented so that $\rho>0$ means higher SNR goes with lower error; number of subjects with a positive $\rho$; two-sided Wilcoxon signed-rank $p$ over subjects. For comparison, the cross-method $\rho$ of the HR pool indicators is 0.25--0.30 on OMS-Test and 0.26--0.30 on MAHNOB-HCI (Table~\ref{tab:s6hr}).\end{minipage}
\end{table}

Over the other method groups (NLM, DLM, DN1), no variant beat the equal mean except one exploratory comparison: the ridge model on all indicators in the NLM group ($+3.8$\% [1.5, 5.7]), which was still no better than the constant predictor ($-6.0$\% [$-12.2$, 0.3]). This is one significant result among 24 group-level comparisons with the equal mean and is not interpreted further.
\FloatBarrier

\end{document}